\documentclass[a4paper,10pt,conference]{ieeeconf}

\pdftrailerid{}
\usepackage{cite}
\usepackage{amsmath,amsfonts}
\usepackage{graphicx}
\usepackage{xcolor}
\usepackage{capt-of}
\usepackage{booktabs}
\usepackage{multirow}
\usepackage{colortbl}
\usepackage{collcell}

\newcommand{\pmark}[1]{\rlap{$^{#1}$}}
\newlength{\resultlitpoint}
\makeatletter

\newcommand{\resultlitcell}[1]{%
  \in@{.}{#1}%
  \ifin@\result@litsplit#1\@nil\else#1\fi}
\def\result@litsplit#1.#2\@nil{%
  \makebox[\resultwidth{mean}][l]{\hspace{\resultlitpoint}\llap{#1}.#2}}
\makeatother
\newcommand{\stdsuffix}[1]{\kern0.1em\raisebox{-0.4ex}{\scriptsize\color{black}$\pm$#1}}
\newcounter{resulttable}
\newif\ifresultmeasure
\newsavebox{\resultmeasurebox}
\newlength{\resultpadding}
\newcommand{\resultkey}[1]{result@\theresulttable @\resultcolumnname @#1}
\newcommand{\resultwidth}[1]{%
  \ifcsname\resultkey{#1}\endcsname\csname\resultkey{#1}\endcsname\else0pt\fi}
\newcommand{\resultmeasurevalue}[2]{%
  \settowidth{\resultpadding}{#2}%
  \ifdim\resultpadding>\resultwidth{#1}\relax
    \expandafter\xdef\csname\resultkey{#1}\endcsname{\the\resultpadding}%
  \fi}
\newcommand{\resultcolumnspace}{%
  \ifresultmeasure\else
    \settowidth{\resultpadding}{\resultcolumnname}%
    \ifdim\resultpadding<\resultwidth{mean}\relax\setlength{\resultpadding}{\resultwidth{mean}}\fi
    \setlength{\resultpadding}{\dimexpr\resultwidth{mean}/2+\resultwidth{sd}-\resultpadding/2\relax}%
    \ifdim\resultpadding>0pt\hspace{\resultpadding}\fi
  \fi}
\ExplSyntaxOn
\regex_const:Nn \c__result_numeric_regex
  { \A (?: \c{.*} | \cB. | \cE. | \cM. | [\-\+] | \s )* [0-9] }
\prg_generate_conditional_variant:Nnn \regex_match:Nn { NV } { TF }
\tl_new:N \l__result_cell_tl
\NewDocumentCommand{\resultcell}{+m}
  {
    \tl_set:Nn \l__result_cell_tl {#1}
    \tl_trim_spaces:N \l__result_cell_tl
    \regex_match:NVTF \c__result_numeric_regex \l__result_cell_tl
      { \exp_args:NV \resultnumber \l__result_cell_tl }
      { \l__result_cell_tl }
  }
\ExplSyntaxOff
\newcommand{\resultcolumncell}[2]{\def\resultcolumnname{#1}\resultcell{#2}}
\newcommand{\resultnumber}[1]{%
  \ifresultmeasure\resultmeasurevalue{mean}{#1}\fi
  \makebox[\resultwidth{mean}][r]{#1}}
\newcolumntype{G}[1]{>{\collectcell{\resultcolumncell{#1}}}c<{\endcollectcell\resultcolumnspace}}
\newcolumntype{N}[1]{>{\collectcell{\resultcolumncell{#1}}}c<{\endcollectcell}}
\DeclareRobustCommand{\cellstd}[2]{\relax%
  \ifdefined\resultcolumnname
    \ifresultmeasure
      \resultmeasurevalue{mean}{#1}%
      \resultmeasurevalue{sd}{\stdsuffix{#2}}%
    \fi
    \makebox[\resultwidth{mean}][r]{#1}\rlap{\stdsuffix{#2}}%
  \else
    \makebox[0.5\width]{\hphantom{\stdsuffix{#2}}}#1\stdsuffix{#2}%
  \fi}
\NewDocumentEnvironment{resulttabular}{o m +b}{%
  \stepcounter{resulttable}%
  \resultmeasuretrue
  \sbox{\resultmeasurebox}{\begin{tabular}{#2}#3\end{tabular}}%
  \resultmeasurefalse
  \IfNoValueTF{#1}%
    {\begin{tabular}{#2}#3\end{tabular}}%
    {\begin{tabular*}{#1}{#2}#3\end{tabular*}}%
}{}
\makeatletter
\newcommand{\continuousresultbackgrounds}{%
  \let\result@paintbackground\CT@@do@color
  \def\CT@@do@color{%
    \pdfsave
    \pdfsetmatrix{1 0 0 1.01}%
    \result@paintbackground
    \pdfrestore}}
\newcommand{\resultcellbackground}[2]{%
  \gdef\CT@cell@color{%
    \CT@color{gray!10}%
    \@tempdimb#1\relax
    \@tempdimc#2\relax
    \global\let\CT@cell@color\relax}}
\makeatother
\DeclareRobustCommand{\indom}[1]{\cellcolor{gray!10}#1}

\usepackage[colorlinks,linkcolor=black,citecolor=black,urlcolor=blue,implicit=false]{hyperref}

\IEEEoverridecommandlockouts
\newcommand{\papertitle}{FlexGaze: Weakly Supervised Gaze Estimation at Scale}
\hypersetup{pdftitle={\papertitle},pdfauthor={Zongjian Li and Mohammad Soleymani},pdfcreator={TeX}}
\title{\LARGE\bfseries \papertitle}
\author{%
  \authorblockN{Zongjian Li \quad Mohammad Soleymani}
  \authorblockA{USC Institute for Creative Technologies\\
    \texttt{\{lizongjian,soleymani\}@ict.usc.edu}}%
}

\begin{document}

\IEEEaftertitletext{\begin{minipage}{\textwidth}
  \centering
  \includegraphics[width=\textwidth]{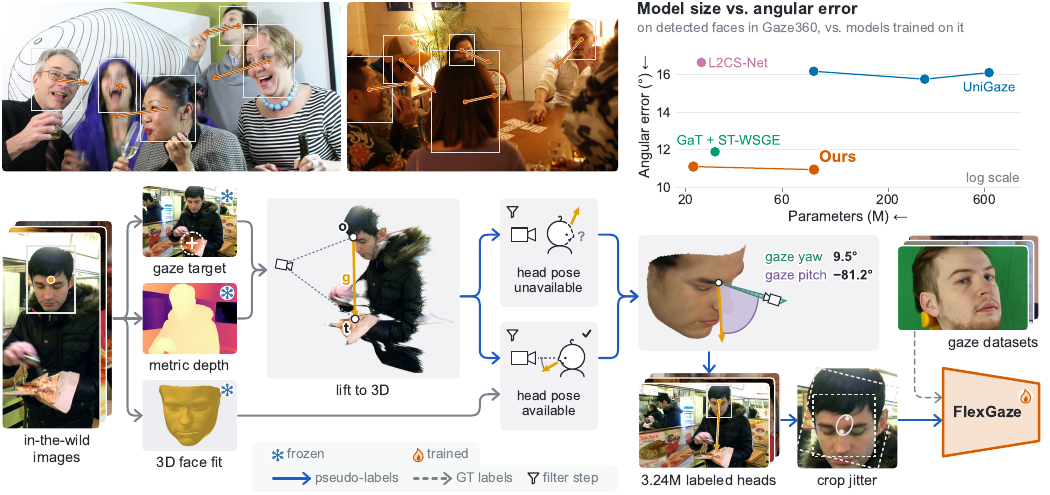}
  \captionof{figure}{Overview of FlexGaze. Top left: example predictions with 3D gaze directions projected onto the image plane. Top right: parameter count versus mean angular error on the face-detected subset of Gaze360. Bottom: the automatic label generation pipeline and model training.}
  \label{fig:teaser}
\end{minipage}
\vspace{4pt}
}
\maketitle
\thispagestyle{empty}

\begin{abstract}
Most gaze estimation models are trained on data collected using specialized equipment. The limited diversity of subjects, scenes, and head poses in these datasets hinders generalization to new conditions. We present FlexGaze, a family of 3D gaze estimators pretrained with weak supervision from gaze labels (pseudo-labels) automatically generated from in-the-wild images. We use off-the-shelf models to predict 2D gaze targets and scene depth, then combine these predictions to generate 3D gaze labels for approximately 3.24 million head instances. At inference, FlexGaze predicts gaze from a head crop alone and can handle profile and rear views without requiring face detection or head pose estimation. We use data augmentation during training to improve robustness to variations in crop position and scale. Pretrained using only these generated gaze labels, FlexGaze generalizes to unseen benchmarks. After supervised fine-tuning on public gaze datasets, we evaluate FlexGaze on eight benchmarks under a unified evaluation protocol. Our models match or outperform publicly available gaze estimators with several times as many parameters on most benchmarks. We release our dataset of generated gaze labels, all trained models, and code via \href{https://github.com/ihp-lab/FlexGaze}{\mbox{github.com/ihp-lab/FlexGaze}}.

\end{abstract}

\section{Introduction}
\label{sec:introduction}

3D gaze estimation infers a person's gaze direction from an image, providing cues to visual attention and interaction~\cite{smith2013gazelocking,hu2023gfie}. In everyday scenes, models must handle variation in subjects, lighting, subject-to-camera distance, and head pose. Obtaining 3D gaze annotations typically requires camera calibration, known fixation targets, or specialized capture equipment~\cite{zhang2020ethxgaze,fischer2018rtgene}, making it costly to collect diverse training data. Input preprocessing presents another challenge. Methods that rely on face localization and normalization~\cite{zhang2017fullface,zhang2018normalization} are limited when faces cannot be reliably localized or are not visible. Differences in detectors and cropping conventions across applications also change crop position and scale, causing inputs to depart from the framing seen during training.

In-the-wild images offer another source of gaze supervision. The relative 3D positions of the head and gaze target determine gaze direction. Advances in gaze target prediction~\cite{ye2026page,ryan2025gazelle,tafasca2024sharingan} and monocular estimation of depth and camera parameters~\cite{piccinelli2026unidepthv2,piccinelli2024unidepth,wang2025moge2,lin2026depthanything3} allow these positions to be inferred from images. We automatically generate 3D gaze labels for weakly supervised pretraining from diverse image corpora, without manual gaze or target annotations for new images. Related work also uses gaze targets. ST-WSGE~\cite{vuillecard2025gat} refines predicted gaze with manually annotated 2D targets, while GeomGaze~\cite{tafasca2023childplay} combines such annotations with estimated scene geometry to construct 3D gaze labels.

We present FlexGaze, a family of 3D gaze estimators that take head crops as input and are pretrained on automatically generated labels (Fig.~\ref{fig:teaser}). To retain training examples with rear views or facial occlusion, where head pose is often difficult to estimate, we use separate filtering branches for samples with and without reliable head pose. After pretraining, we fine-tune with ground-truth gaze labels (Fig.~\ref{fig:finetuning-inference}) to reduce the effects of errors in the generated labels. During training, we jointly vary crop center and scale to improve robustness to differences in cropping across detectors and applications. At inference, FlexGaze requires neither face detection nor head pose estimation.

Pretraining on generated labels alone enables FlexGaze to generalize across datasets. We compare FlexGaze with UniGaze~\cite{qin2026unigaze}, a representative general-purpose gaze estimator, when both are fine-tuned on the same five datasets. FlexGaze-B achieves lower mean angular errors than UniGaze-H on six of eight benchmarks with approximately one seventh as many parameters. In crop sensitivity experiments, FlexGaze-B also shows flatter error curves overall than the other evaluated models.

Our contributions are threefold:
\begin{itemize}
  \item We introduce a method for automatically constructing 3D gaze supervision by combining predicted 2D gaze targets with monocular scene geometry. It can label new images without manual gaze or target annotations, even when faces are not visible.
  \item We present the FlexGaze model family and a training method based on generated supervision. These compact models achieve competitive accuracy from head crops, including when faces are not visible, and accommodate variations in crop position and scale.
  \item We construct a dataset of generated 3D gaze labels from $17$ public image corpora, retaining approximately $3.242$ million head instances after filtering, more than twice the $1.4$ million faces pseudo-labeled by OmniGaze~\cite{qu2025omnigaze}. It covers diverse gaze directions and imaging conditions.
\end{itemize}
We include the unfiltered annotations and metadata in the public release so users can apply their own filters. The release also includes model weights, training and inference code, and a reusable evaluation toolkit.

\section{Related Work}
\label{sec:related-work}

\paragraph{Gaze Estimation and Generalization}

Appearance-based 3D gaze estimation predicts gaze direction from images of the eyes, face, or head. Full-face models~\cite{zhang2017fullface}, GazeTR~\cite{cheng2022gazetr}, and L2CS-Net~\cite{abdelrahman2023l2cs} use facial appearance, with some methods normalizing their inputs using landmarks, head pose, and camera parameters~\cite{zhang2018normalization}. The Gaze360 baseline~\cite{kellnhofer2019gaze360} and GaT~\cite{vuillecard2025gat} instead use head crops, extending prediction beyond frontal faces. When faces are not visible, other approaches infer gaze from head and body orientations over time~\cite{nonaka2022dynamic} or from 3D human pose and scene objects~\cite{kawana2025ga3ce}, but both rely on video or 3D scene information. Among gaze datasets, GazeCapture~\cite{krafka2016eyetracking} broadens subject diversity, ETH-XGaze~\cite{zhang2020ethxgaze} covers wider ranges of head poses and gaze angles, and Gaze360 includes indoor and outdoor scenes as well as rear views. Further expanding this coverage still requires gaze measurement.

Synthetic data, representation learning, and domain adaptation offer other ways to improve generalization. UnityEyes~\cite{wood2016unityeyes} renders eye images with varied appearance, lighting, and gaze direction. PureGaze~\cite{cheng2022puregaze} uses self-adversarial learning to suppress nuisance factors such as identity and illumination, while Xu et al.~\cite{xu2023gazeconsistent} learn gaze-consistent features by perturbing factors unrelated to gaze. UniGaze~\cite{qin2026unigaze} performs masked autoencoding on a large collection of normalized face images before learning gaze prediction from annotated data. To reduce domain shift, other methods adapt the model with unlabeled target-domain images~\cite{liu2021pnpga,wang2022contrastive,bao2022rotation,cai2023sourcefree} or align gaze label distributions across datasets~\cite{zeng2025labelalignment}. These representation learning and adaptation methods still learn gaze from existing annotations, whereas our approach expands the supervision data and can complement them.

\paragraph{Weak Supervision for Gaze Estimation}

Weak supervision makes images without measured gaze annotations usable for training 3D gaze estimators. Kothari et al.~\cite{kothari2021weakly} derive geometric constraints from mutual gaze annotations, reducing the need for ground-truth 3D directions. ST-WSGE~\cite{vuillecard2025gat} uses manually annotated 2D targets~\cite{recasens2015where}, including those for people facing away from the camera, to refine predictions from a model trained on measured gaze data. GaT is then trained jointly on ground-truth and refined gaze labels. 3DGazeNet~\cite{ververas2024gazenet} generates pseudo-labels through 3D face alignment, iris localization, and eyeball fitting, and adds consistency supervision across synthetic views. OmniGaze~\cite{qu2025omnigaze} labels unlabeled face images with a model trained on measured gaze data and learns a reward model to select and weight the labels. Both methods primarily generate labels for visible faces.

\paragraph{Gaze Target Estimation and Scene Geometry}

Gaze target estimation combines a person's appearance with scene context to locate where they are looking in an image~\cite{recasens2015where,chong2020attended}. Recent methods predict targets for multiple people jointly~\cite{tafasca2024sharingan} or use large pretrained visual encoders for target prediction~\cite{ryan2025gazelle}. PaGE~\cite{ye2026page} uses interactions between head and scene features to improve target localization. GFIE~\cite{hu2023gfie} provides both 2D and 3D target annotations, extending the task to 3D scenes. The 3D positions of the gaze origin and target provide direction supervision. For gaze target localization, GeomGaze, introduced with ChildPlay~\cite{tafasca2023childplay}, constructs 3D gaze labels from manually annotated 2D targets and estimated point clouds. Its head-crop gaze branch uses a Gaze360-pretrained backbone and is trained jointly with a scene branch. We automatically generate labels from predicted 2D targets and estimated scene geometry to pretrain FlexGaze, a standalone head-crop gaze estimator evaluated on measured gaze benchmarks. Scene information is used only for label generation.

\section{Method}
\label{sec:method}

\subsection{Automatic 3D Gaze Supervision}
\label{sec:method:pipeline}
\label{sec:method:filtering}

We construct 3D gaze labels from in-the-wild images by combining predictions of 2D gaze targets and scene geometry. As illustrated in Fig.~\ref{fig:teaser}, these labels supervise FlexGaze, which estimates gaze from a head crop. To identify the head instances, we apply YOLOv5m~\cite{jocher2021yolov5} pretrained on CrowdHuman~\cite{shao2018crowdhuman} to each image at multiple resolutions and consolidate the detections by removing low-confidence, duplicate, and nested boxes.

\begin{figure}[t]
  \centering
  \includegraphics[width=\columnwidth]{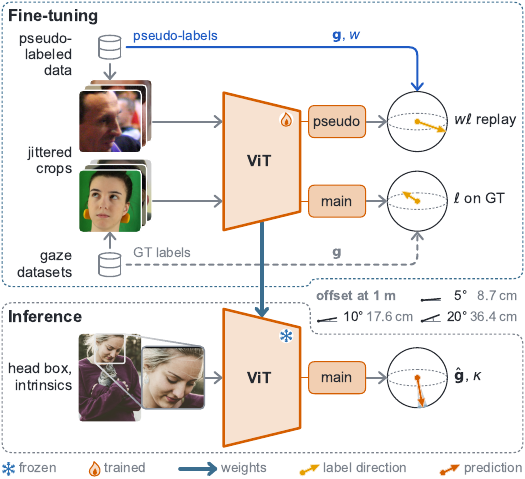}
  \caption{Fine-tuning and inference of FlexGaze. Top: a main head learns from ground-truth labels and a pseudo-label head from replayed pseudo-labels. Bottom: only the main head is used at inference.}
  \label{fig:finetuning-inference}
\end{figure}

For each detected head, PaGE~\cite{ye2026page} takes the scene image and head box as input and predicts a gaze target heatmap. We obtain the 2D target from the probability-weighted centroid around the heatmap peak. We use UniDepthV2~\cite{piccinelli2026unidepthv2,piccinelli2024unidepth} to estimate a depth map and camera intrinsics for the image. Using the depths at the head box center and predicted target, we back-project the two points into camera coordinates. We take the resulting 3D points as the gaze origin $\mathbf{o}$ and target $\mathbf{t}$, respectively, and define the unit gaze label as ${\mathbf{g}=(\mathbf{t}-\mathbf{o})/\|\mathbf{t}-\mathbf{o}\|_2}$.

We refine the generated labels through a fixed target depth correction and filtering. Basic checks remove samples with small head boxes, low-confidence targets, invalid depths, or implausible physical spans estimated from the head boxes and scene geometry. We then use one of two filtering branches, depending on whether head pose is available. We estimate pose through 3D face fitting with the 3DDFA-V3 pipeline~\cite{wang2024geometric,lei2023hrn,deng2020retinaface} and match the fits to the detected heads. For heads with matched fits, we constrain gaze deviation relative to the head using bounds informed by eye rotation ranges~\cite{guitton1987gaze,lee2019eyemovement,gerling1997duction}, with allowance for pose estimation errors~\cite{martyniuk2022dad3dheads}. When head pose is unavailable, we exclude generated gaze labels pointing approximately toward the camera, because such views usually show a face that the fitting should recover. This branch still retains training samples with extreme head poses or facial occlusion. The correction and filtering settings are specified in Sec.~\ref{sec:experiments:setup}.

\subsection{Input Representation and Data Augmentation}
\label{sec:method:representation}

FlexGaze uses head boxes to accommodate profile and rear views as well as occluded or partially visible faces. The head box center serves as both the default viewing center and the origin of the generated gaze label. Following virtual-camera-based normalization~\cite{zhang2018normalization}, we use the camera intrinsics to orient a virtual camera toward this center, with the original camera's horizontal axis as the reference for image orientation. The shorter side of the head box projected into this view defines a square crop. At inference, this input representation requires no face detection, facial landmarks, head pose fitting, or depth estimation.

During training, we jointly vary the crop center and scale to accommodate differences in input boxes across detectors and applications. Because the eyes may lie near the head box boundary in profile views, large center shifts or tight crops can exclude them. We therefore use a face center~\cite{zhang2017fullface} derived from facial landmarks as an auxiliary reference, which allows tighter crops around the face and improves compatibility with face-box inputs. We sample crop centers within an ellipse whose foci are the projected head box center and face center. We then randomly rescale the window within a range that widens continuously from the head box center toward the face center, where the eyes are more likely to remain inside tighter crops. When the face center is unavailable, we sample crop centers within a circle around the head box center and use the scale range defined at that center. We refer to these joint center and scale variations as crop jitter. %

Changing the crop center reorients the virtual camera but leaves the original 3D gaze origin unchanged. Our crop augmentation also includes random camera roll to vary image orientation. %
The roll is composed with the normalization rotation before sampling the augmented view from the original image, and the same composed rotation transforms the gaze label. We also apply horizontal flipping, color jitter, Gaussian blur, and resolution degradation. Flipping negates the label's horizontal component, whereas the appearance augmentations leave it unchanged.

\subsection{Gaze Estimation and Learning}
\label{sec:method:model}

FlexGaze consists of a ViT backbone~\cite{dosovitskiy2021vit} initialized with DINOv3 weights~\cite{simeoni2025dinov3} and a gaze prediction head, trained jointly. The head predicts a unit gaze direction $\hat{\mathbf{g}}\in\mathbb{R}^3$ in the virtual camera coordinate system, together with a concentration $\kappa>0$. We model gaze with a von Mises--Fisher (vMF) distribution~\cite{fisher1953dispersion} on the unit sphere, where $\hat{\mathbf{g}}$ is the mean direction and $\kappa$ controls concentration around the predicted direction. For sample $i$ with a unit gaze label $\mathbf{g}_i\in\mathbb{R}^3$ expressed in the virtual camera coordinate system, the negative log-likelihood is
\begin{equation}
  \ell_i=-\log C(\kappa_i)
  -\kappa_i\hat{\mathbf{g}}_i^\top\mathbf{g}_i,
  \label{eq:vmf-sample-loss}
\end{equation}
where $C(\kappa)=\kappa/(4\pi\sinh\kappa)$ makes the density integrate to one over the sphere, and the dot product measures agreement between the predicted and labeled directions. Both direction and concentration are learned from the input through this objective. To account for differences in label confidence, we form a score by multiplying PaGE's predicted probability that the gaze target lies inside the scene image by the peak probability of its target heatmap. Dividing each score by the batch mean gives sample weights $w_i$ with a mean of one. Pretraining minimizes the batch mean of $w_i\ell_i$. At inference, we transform $\hat{\mathbf{g}}$ back to the original camera coordinate system using the inverse normalization rotation.

The pretrained model can be used directly. We also fine-tune it with ground-truth gaze labels to reduce the effects of errors and systematic biases that generated labels may inherit from target prediction and depth estimation. However, data collected for screen interaction or controlled laboratory settings~\cite{zhang2019mpiigaze,krafka2016eyetracking,zhang2020ethxgaze} differ in input distribution from the in-the-wild pretraining images. Fine-tuning can reduce generalization under such distribution shifts~\cite{kumar2022finetuning,wortsman2022robust}. To retain supervision on the pretraining distribution during adaptation, we use pseudo-labeled data replay, mixing pretraining images and their generated labels with ground-truth-labeled data. Two separate prediction heads share the backbone and receive ground-truth and pseudo-label supervision, respectively (Fig.~\ref{fig:finetuning-inference}). This allows the head retained for inference to learn from ground-truth gaze labels without directly fitting potential pseudo-label biases. Both losses update the shared backbone. Using the outputs of the corresponding head in Eq.~\ref{eq:vmf-sample-loss}, we optimize
\begin{equation}
  \mathcal{L}_{\mathrm{ft}}=
  \frac{1}{|\mathcal{B}|}
  \left(
    \sum_{i\in\mathcal{B}_{\mathrm{gt}}}\ell_i
    +\sum_{i\in\mathcal{B}_{\mathrm{p}}}w_i\ell_i
  \right),
  \label{eq:finetuning-loss}
\end{equation}
where $\mathcal{B}_{\mathrm{gt}}$ and $\mathcal{B}_{\mathrm{p}}$ partition the mixed batch $\mathcal{B}$ into ground-truth-labeled and pseudo-labeled samples. The weights $w_i$ follow the pretraining definition but are normalized within $\mathcal{B}_{\mathrm{p}}$. Inference uses only the shared backbone and the head supervised by ground-truth gaze labels.

\section{Experiments}
\label{sec:experiments}

\subsection{Experimental Setup}
\label{sec:experiments:setup}

We assemble a pretraining corpus of real images from $17$ public datasets, favoring those that retain scene context for gaze target localization and 3D geometry estimation. The sources are Open Images V7~\cite{kuznetsova2020openimages}, IMDB-WIKI~\cite{rothe2018imdbwiki,rothe2015dex}, GazeFollow~\cite{recasens2015where}, LAGENDA~\cite{kuprashevich2024mivolo,kuprashevich2025beyond}, ExpW~\cite{zhang2018expw}, WIDER FACE~\cite{yang2016widerface}, GAF 3.0~\cite{dhall2018emotiw}, Flickr30k~\cite{young2014flickr30k}, VideoAttentionTarget (VAT)~\cite{chong2020attended}, MPII Human Pose~\cite{andriluka2014mpiihumanpose}, FFHQ~\cite{karras2019stylegan}, FairFace~\cite{karkkainen2021fairface}, CelebA~\cite{liu2015celeba}, HECO~\cite{yang2022heco}, MERL-RAV~\cite{kumar2020luvli,koestinger2011aflw}, LaPa~\cite{liu2020lapa}, and Menpo2D~\cite{deng2019menpo,zafeiriou2017menpo}. We use the dataset images but not their original task annotations. GazeFollow and VAT are restricted to their training splits, while all images from the other selected releases are included. Image deduplication using pixel hashes and perceptual hashes (pHash)~\cite{zauner2011rihamark}, followed by pseudo-label quality filtering, yields approximately $3.242$ million head instances with generated 3D gaze labels.

\begin{figure*}[t]
  \centering
  \includegraphics[width=\textwidth]{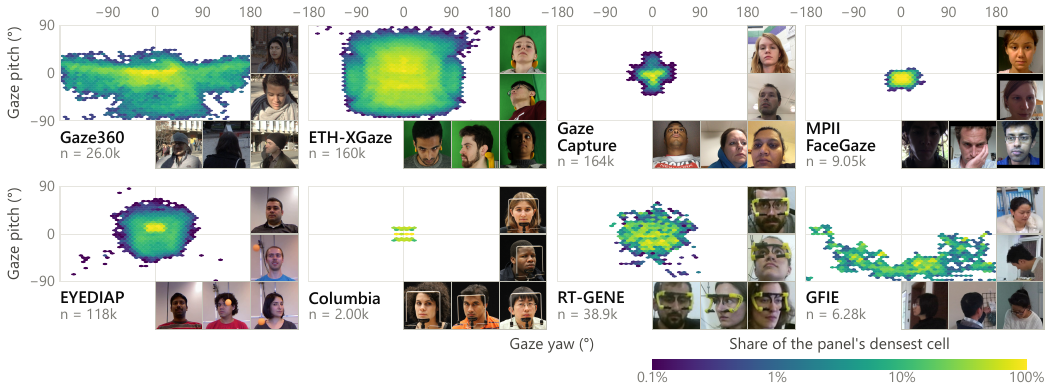}
  \caption{Sample images from eight gaze datasets and ground-truth gaze angle distributions on their test splits. Zero yaw and pitch indicate gaze toward the camera. Heatmap counts are normalized by the peak count in each panel. $n$ denotes the number of test samples included in each angular distribution.}
  \label{fig:gaze-dataset-overview}
\end{figure*}

\begin{figure}[t]
  \centering
  \includegraphics[width=\columnwidth]{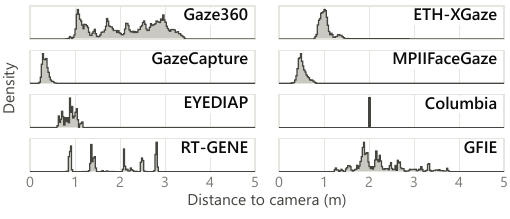}
  \caption{Subject-to-camera distance distributions on the test splits of eight gaze datasets. GazeCapture distances are estimated using FAZE preprocessing~\cite{park2019faze}.}
  \label{fig:gaze-dataset-distances}
\end{figure}

For supervised fine-tuning and evaluation, we use the training and test splits of eight datasets with measured gaze annotations: Gaze360~\cite{kellnhofer2019gaze360}, ETH-XGaze~\cite{zhang2020ethxgaze,qin2025xgaze3d}, GazeCapture~\cite{krafka2016eyetracking}, MPIIFaceGaze~\cite{zhang2017fullface}, EYEDIAP~\cite{funesmora2014eyediap}, RT-GENE~\cite{fischer2018rtgene}, Columbia~\cite{smith2013gazelocking}, and GFIE~\cite{hu2023gfie}. MPIIFaceGaze, EYEDIAP, Columbia, and RT-GENE use $10/5$, $8/8$, $37/19$, and $10/5$ training/test subject splits, the first two following UniGaze~\cite{qin2026unigaze}, and EYEDIAP is tested on every $15$th frame. Figs.~\ref{fig:gaze-dataset-overview} and~\ref{fig:gaze-dataset-distances} show example images and test-set distributions of ground-truth gaze angles and subject-to-camera distances.

For label generation, we set the filtering thresholds empirically through visual inspection of geometric reconstructions. Basic checks require a head box short side of at least $16$ pixels, PaGE in-frame and heatmap peak probabilities of at least $0.5$ and $0.2$, respectively, and valid raw depths of at least $0.1$ m at the head center and target. The larger physical head box dimension, estimated from head depth and camera intrinsics, must lie between $35$ and $1600$ mm. We scale target depth by $e^{-0.03}$, a factor chosen based on results from the GFIE validation split. We require a 3D separation of at least $0.05$ m between the head center and the corrected target. When head pose is available, the horizontal and vertical components of the total gaze deviation from the head's forward direction must satisfy an elliptical bound with horizontal, upward, and downward semiaxes of $60^\circ$, $60^\circ$, and $80^\circ$. Otherwise, gaze must deviate by more than $30^\circ$ from the head-to-camera direction.

FlexGaze-S and FlexGaze-B use ViT-S/16 and ViT-B/16 backbones, respectively, with $224\times224$ inputs. We initialize the backbones with DINOv3 weights and pretrain S and B for $30$ and $20$ epochs, respectively. We average epoch checkpoints from the second half of each stage with equal weights, separately for each seed. Each model is then fine-tuned for $10$ epochs from the averaged weights for the same size and seed. Fine-tuning uses separate prediction heads, with each epoch combining all training samples with ground-truth gaze labels and an equal number of pretraining samples for replay. In both stages, AdamW~\cite{loshchilov2019adamw} jointly optimizes the backbone and gaze prediction heads at fixed learning rates of $2\times10^{-5}$ for the backbone and $2\times10^{-4}$ for the heads. Training uses NVIDIA H100 80 GB GPUs with a global effective batch size of $4096$. Both stages sample camera roll within $\pm15^\circ$ and crop centers within an ellipse extending beyond each focus by $0.1$ times the shorter head box side, with crop scale ranges of $[0.9,1.1]$ at the head box center and $[0.5,1.5]$ at the face center. Main results report the mean and sample standard deviation across independent runs with seeds $0,1,2$.

\subsection{Quality and Coverage of Generated Supervision}
\label{sec:experiments:supervision}

Using the scene images and measured gaze annotations of GFIE, we assess label quality on its training split, which is not used to select the depth correction factor (Table~\ref{tab:gfie-label-quality}). The generated labels have a mean angular error of $15.93^\circ$, compared with $37.68^\circ$ for a constant prediction of the ground-truth mean direction, indicating that they provide gaze information beyond the dataset's overall directional bias. The mean 2D target localization error is $48.01$ pixels. To isolate depth error, we compare raw predicted and measured depths at the annotated target pixels and find a higher mean absolute error (MAE) than at the head box centers, with target depths overestimated on average. Together, these results indicate that both target localization and depth estimation errors affect the accuracy of the generated gaze labels.

\begin{table}[t]
  \centering
  \caption{Label generation errors on the GFIE training split.}
  \label{tab:gfie-label-quality}
  \begingroup
  \footnotesize
  \setlength{\tabcolsep}{1.3pt}
  \renewcommand{\arraystretch}{1.12}
  \begin{tabular*}{\columnwidth}{@{}l@{\hspace{2pt}}r@{\extracolsep{\fill}\hspace{2\tabcolsep}}rrrrr@{}}
    \toprule
    \multicolumn{1}{@{}c@{\hspace{2pt}}}{\multirow{2}{*}{GFIE samples}} & \multirow{2.7}{*}[2pt]{\shortstack{Target\\2D error\\(px)}} & \multicolumn{2}{c}{Head depth (m)} & \multicolumn{2}{c}{Target depth (m)} & \multicolumn{1}{c}{\multirow{2.7}{*}[2pt]{\shortstack{Gaze\\error\\($^{\circ}$)}}} \\
    \cmidrule(lr){3-4}\cmidrule(lr){5-6}
    & & \multicolumn{1}{c}{MAE} & \multicolumn{1}{c}{$\overline{\Delta}$} & \multicolumn{1}{c}{MAE} & \multicolumn{1}{c}{$\overline{\Delta}$} & \\
    \midrule
    Before filtering & 48.40 & $0.127$ & $+0.011$ & $0.202$ & $+0.152$ & 18.70 \\
    \hspace{0.6em}Pose available & 48.04 & $0.124$ & $+0.002$ & $0.193$ & $+0.140$ & 18.74 \\
    \hspace{0.6em}Pose unavailable & 52.38 & $0.170$ & $+0.148$ & $0.298$ & $+0.279$ & 18.20 \\
    \midrule
    After filtering & 48.01 & $0.128$ & $+0.008$ & $0.195$ & $+0.144$ & 15.93 \\
    \bottomrule
  \end{tabular*}
  \par\vspace{3pt}
  \begin{minipage}{\columnwidth}
    \footnotesize\raggedright
    \par
    $\overline{\Delta}$ denotes the mean signed depth error (prediction minus measurement).
  \end{minipage}
  \endgroup
\end{table}

\begin{figure}[t]
  \centering
  \includegraphics[width=\columnwidth]{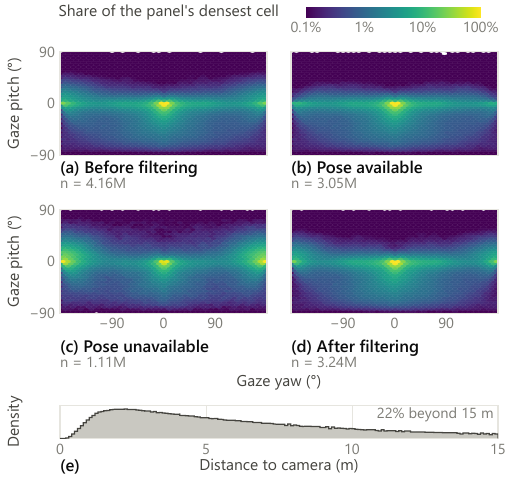}
  \caption{Distributions of generated gaze angles and estimated subject-to-camera distances in the pretraining corpus. (a) All candidates before quality filtering, (b,c) those with and without head pose estimates, (d) samples retained for pretraining, and (e) distances of the retained samples. Angle conventions and heatmap normalization follow Fig.~\ref{fig:gaze-dataset-overview}, and $n$ is the number of head instances.}
  \label{fig:pseudo-label-angle-coverage}
\end{figure}

Figure~\ref{fig:pseudo-label-angle-coverage} shows the gaze angle distributions across the full pretraining corpus and the estimated subject-to-camera distance distribution after filtering. Compared with most gaze benchmarks in Figs.~\ref{fig:gaze-dataset-overview} and~\ref{fig:gaze-dataset-distances}, the generated labels cover a wider range of gaze angles. The estimated distances also follow a relatively continuous distribution over a broad range, providing pretraining examples at different distances. To include samples without reliable head pose in pretraining, our supervision pipeline uses separate filtering branches depending on whether head pose is available. Their angular distributions before filtering (Fig.~\ref{fig:pseudo-label-angle-coverage}(b,c)) show that samples with reliable pose are concentrated around gaze directions toward the camera. Large gaze yaw angles, especially directions away from the camera, are more prominent among samples without reliable pose. Retaining this branch therefore broadens the directional coverage of pretraining with large-angle gaze samples, which are scarce in most gaze benchmarks (Fig.~\ref{fig:gaze-dataset-overview}).

\subsection{Comparison Protocols}
\label{sec:experiments:comparison}

We compare FlexGaze with 3D estimators trained with ground-truth gaze labels, using their publicly released weights under a unified evaluation protocol: GaT + ST-WSGE~\cite{vuillecard2025gat}, L2CS-Net~\cite{abdelrahman2023l2cs}, PureGaze~\cite{cheng2022puregaze}, the ETH-XGaze baseline~\cite{zhang2020ethxgaze}, GazeTR-Hybrid~\cite{cheng2022gazetr}, and UniGaze-B/L/H~\cite{qin2026unigaze}. Throughout the tables, gray shading marks in-domain results, where training used ground-truth gaze labels from the evaluated dataset.

We preserve each method's training-time crop extents, normalization parameters, and geometric point definitions, generally following its released detection and cropping pipeline. Methods trained on dataset-provided crops or normalized images use those images for evaluation. We obtain facial landmarks, head pose, facial reference points, and camera intrinsics from shared dataset annotations and official preprocessing outputs where available. When landmarks must be estimated, we derive geometry with STAR~\cite{zhou2023star} and FAN~\cite{bulat2017facealignment} and report the lower mean angular error for each competing method on each dataset.

In each comparison, we report mean angular error on the test samples for which every participating model can complete input preprocessing and gaze prediction. Gaze360 and GFIE include rear views and other samples with invisible faces, so we divide their test sets based on face detection and landmark localization. We use SCRFD~\cite{guo2021scrfd} to detect the subject's face. The face-detected subset also requires successful detection by RetinaFace~\cite{deng2020retinaface}, which is used in L2CS-Net's official pipeline, and FAN landmarks that support valid eye crops. Samples with failed face detection or landmark localization form the face-undetected subset, where we evaluate methods that do not require face detection.

Because methods define different gaze origins (head box center, face center, or midpoint between the eyes), we place a point along each predicted direction at the ground-truth target distance and evaluate the direction from the dataset-defined origin to this point. For RT-GENE, which lacks target positions, directions are compared without this adjustment.

\subsection{Main Comparison and Supervised Fine-Tuning}
\label{sec:experiments:finetuning}

After joint fine-tuning on all eight gaze datasets, FlexGaze-B outperforms the compared methods on five of the eight benchmarks (Table~\ref{tab:face-detected-results}). Compared with UniGaze, one of the leading general-purpose gaze estimation methods, FlexGaze-B uses $86.0$M parameters at inference, approximately one seventh as many as UniGaze-H, while achieving lower errors on most benchmarks. FlexGaze-S uses only $21.8$M parameters, approximately one quarter as many as UniGaze-B, yet outperforms it on five of the eight benchmarks. FlexGaze achieves these results with simpler input preprocessing, requiring only a head box and camera intrinsics, without face detection or head pose estimation.

We also fine-tune FlexGaze on the same five datasets used by UniGaze (Table~\ref{tab:face-detected-results}). Under this setting, FlexGaze-B still achieves lower mean angular errors than UniGaze-H on six of the eight test benchmarks, while FlexGaze-S outperforms UniGaze-B on five.

\begin{table*}[t]
  \centering
  \setlength{\tabcolsep}{0.75pt}
  \renewcommand{\arraystretch}{0.94}
  \continuousresultbackgrounds
  \DeclareRobustCommand{\indom}[1]{%
    \resultcellbackground{\dimexpr\tabcolsep+0.05pt\relax}{\dimexpr\tabcolsep+0.05pt\relax}#1}
  \newlength{\mainresultleft}
  \newlength{\mainresultmeanleft}
  \newlength{\mainresultright}
  \newlength{\mainresultwidth}
  \newlength{\mainresultshift}
  \DeclareRobustCommand{\mainresultheader}[1]{%
    \ifresultmeasure
      \resultmeasurevalue{header}{\begin{tabular}[c]{@{}c@{}}#1\end{tabular}}%
    \fi
    \begin{tabular}[c]{@{}c@{}}#1\end{tabular}}
  \newcommand{\mainresultsizes}[1]{%
    \def\resultcolumnname{#1}%
    \setlength{\resultpadding}{\dimexpr\resultwidth{header}/2\relax}%
    \ifdim\resultpadding>\mainresultleft\setlength{\mainresultleft}{\resultpadding}\fi
    \ifdim\resultpadding>\mainresultright\setlength{\mainresultright}{\resultpadding}\fi
    \setlength{\resultpadding}{\dimexpr\resultwidth{mean}/2\relax}%
    \ifdim\resultpadding>\mainresultmeanleft\setlength{\mainresultmeanleft}{\resultpadding}\fi
    \ifdim\resultpadding>\mainresultleft\setlength{\mainresultleft}{\resultpadding}\fi
    \addtolength{\resultpadding}{\resultwidth{sd}}%
    \ifdim\resultpadding>\mainresultright\setlength{\mainresultright}{\resultpadding}\fi}
  \renewcommand{\resultcolumnspace}{%
    \ifresultmeasure\else\kern\mainresultshift\fi}
  \newcommand{\mainresultcolumncell}[2]{%
    \makebox[\mainresultwidth][c]{\resultcolumncell{#1}{#2}\resultcolumnspace}}
  \newcolumntype{J}[1]{>{\collectcell{\mainresultcolumncell{#1}}}c<{\endcollectcell}}
  \RenewDocumentEnvironment{resulttabular}{m +b}{%
    \stepcounter{resulttable}%
    \resultmeasuretrue
    \sbox{\resultmeasurebox}{\begin{tabular}{#1}#2\end{tabular}}%
    \mainresultsizes{Gaze360}\mainresultsizes{XGaze}%
    \mainresultsizes{GazeCapture}\mainresultsizes{MPIIFaceGaze}%
    \mainresultsizes{EYEDIAP}\mainresultsizes{Columbia}%
    \mainresultsizes{RT-GENE}\mainresultsizes{GFIE}%
    \addtolength{\mainresultleft}{-3pt}%
    \ifdim\mainresultleft<\mainresultmeanleft
      \setlength{\mainresultleft}{\mainresultmeanleft}%
    \fi
    \setlength{\mainresultwidth}{\dimexpr\mainresultleft+\mainresultright\relax}%
    \setlength{\mainresultshift}{\dimexpr\mainresultright-\mainresultleft\relax}%
    \resultmeasurefalse
    \begin{tabular}{#1}#2\end{tabular}%
  }{}
  \caption{Comparison of FlexGaze with publicly available gaze estimators (mean angular error in degrees, lower is better).}
  \label{tab:face-detected-results}
  \begin{resulttabular}{@{}l!{\hspace{8pt}}N{Params}!{\hspace{8pt}}c!{\hspace{8pt}}J{Gaze360}J{XGaze}J{GazeCapture}J{MPIIFaceGaze}J{EYEDIAP}J{Columbia}J{RT-GENE}J{GFIE}@{}}
    \toprule[0.8pt]
    Method & Params & Preproc.
      & \mainresultheader{Gaze360\\{\scriptsize\upshape\color{black}Face (88\%)}} & \mainresultheader{XGaze}
      & \mainresultheader{Gaze\\Capture} & \mainresultheader{MPII\\FaceGaze}
      & \mainresultheader{EYEDIAP} & \mainresultheader{Columbia}
      & \mainresultheader{RT-GENE\\{\scriptsize\upshape\color{black}Original}} & \mainresultheader{GFIE\\{\scriptsize\upshape\color{black}Face (86\%)}} \\
    \midrule[0.6pt]
    \multicolumn{11}{l}{\textbf{Oracle baseline with constant predictions}} \\
    Mean direction
      & ---
      & ---
      & 51.48 %
      & 42.64 %
      & 7.78 %
      & 11.75 %
      & 23.83 %
      & 12.05 %
      & 27.18 %
      & 36.89 %
      \\
    \midrule[0.4pt]
    \multicolumn{11}{l}{\textbf{Baselines with head-crop input}} \\
    GaT + ST-WSGE
      & 27.9M %
      & H
      & \indom{11.90} %
      & 14.36 %
      & 6.95 %
      & 6.79 %
      & 8.56 %
      & 8.07 %
      & 9.54 %
      & 15.06 %
      \\
    \midrule[0.4pt]
    \multicolumn{11}{l}{\textbf{Baselines with face-crop input}} \\
    L2CS-Net
      & 23.9M %
      & F
      & \indom{16.66} %
      & 17.52 %
      & 10.20 %
      & 9.01 %
      & 11.62 %
      & 8.20 %
      & 30.22 %
      & 38.32 %
      \\
    PureGaze
      & 23.5M %
      & F,L,P,K
      & 33.32 %
      & \indom{4.62} %
      & 7.69 %
      & 9.21 %
      & 8.31 %
      & 7.65 %
      & 18.41 %
      & 57.79 %
      \\
    XGaze baseline
      & 23.5M %
      & F,L,P,K
      & 34.97 %
      & \indom{4.44} %
      & 8.50 %
      & 9.07 %
      & 10.03 %
      & 9.19 %
      & 17.24 %
      & 60.90 %
      \\
    GazeTR-Hybrid
      & 11.4M %
      & F,L,P,K
      & 34.15 %
      & \indom{\textbf{4.29}} %
      & 10.66 %
      & 10.06 %
      & 13.07 %
      & 9.53 %
      & 20.15 %
      & 49.43 %
      \\
    UniGaze-B
      & 85.8M %
      & F,L,P,K
      & \indom{16.19} %
      & \indom{5.73} %
      & \indom{3.83} %
      & \indom{5.80} %
      & \indom{5.16} %
      & 4.56 %
      & 11.00 %
      & 31.96 %
      \\
    UniGaze-L
      & 303.3M %
      & F,L,P,K
      & \indom{15.77} %
      & \indom{5.61} %
      & \indom{3.55} %
      & \indom{5.30} %
      & \indom{4.98} %
      & \textbf{4.22} %
      & 15.97 %
      & 30.47 %
      \\
    UniGaze-H
      & 630.8M %
      & F,L,P,K
      & \indom{16.12} %
      & \indom{5.40} %
      & \indom{3.50} %
      & \indom{5.09} %
      & \indom{\textbf{4.83}} %
      & 4.46 %
      & 26.87 %
      & 30.92 %
      \\
    \midrule[0.6pt]
    \multicolumn{11}{l}{\textbf{Ours, head-crop input, fine-tuned (8 sets)}} \\
    FlexGaze-S
      & 21.8M %
      & H,K
      & \indom{\cellstd{11.11}{0.08}} %
      & \indom{\cellstd{5.97}{0.04}} %
      & \indom{\cellstd{3.68}{0.03}} %
      & \indom{\cellstd{5.35}{0.16}} %
      & \indom{\cellstd{5.86}{0.10}} %
      & \indom{\cellstd{5.01}{0.11}} %
      & \indom{\cellstd{6.86}{0.20}} %
      & \indom{\cellstd{12.17}{0.24}} %
      \\
    FlexGaze-B
      & 86.0M %
      & H,K
      & \indom{\cellstd{10.95}{0.10}} %
      & \indom{\cellstd{5.31}{0.06}} %
      & \indom{\cellstd{\textbf{3.46}}{0.03}} %
      & \indom{\cellstd{\textbf{4.92}}{0.25}} %
      & \indom{\cellstd{5.39}{0.01}} %
      & \indom{\cellstd{4.42}{0.15}} %
      & \indom{\cellstd{\textbf{6.20}}{0.10}} %
      & \indom{\cellstd{\textbf{11.09}}{0.54}} %
      \\
    \midrule[0.4pt]
    \multicolumn{11}{l}{\textbf{Ours, head-crop input, fine-tuned (5 sets shared with UniGaze)}} \\
    FlexGaze-S
      & 21.8M %
      & H,K
      & \indom{\cellstd{11.17}{0.07}} %
      & \indom{\cellstd{5.98}{0.04}} %
      & \indom{\cellstd{3.67}{0.05}} %
      & \indom{\cellstd{5.54}{0.15}} %
      & \indom{\cellstd{5.90}{0.06}} %
      & \cellstd{5.54}{0.42} %
      & \cellstd{10.41}{0.66} %
      & \cellstd{17.70}{1.10} %
      \\
    FlexGaze-B
      & 86.0M %
      & H,K
      & \indom{\cellstd{\textbf{10.90}}{0.15}} %
      & \indom{\cellstd{5.30}{0.03}} %
      & \indom{\cellstd{\textbf{3.46}}{0.01}} %
      & \indom{\cellstd{4.93}{0.21}} %
      & \indom{\cellstd{5.33}{0.05}} %
      & \cellstd{4.52}{0.14} %
      & \cellstd{10.21}{0.25} %
      & \cellstd{14.34}{0.61} %
      \\
    \midrule[0.4pt]
    \multicolumn{11}{l}{\textbf{Ours, head-crop input, pretrained (weak labels only)}} \\
    FlexGaze-S
      & 21.8M %
      & H,K
      & \cellstd{12.69}{0.03} %
      & \cellstd{12.90}{0.07} %
      & \cellstd{10.59}{0.26} %
      & \cellstd{9.97}{0.26} %
      & \cellstd{8.35}{0.27} %
      & \cellstd{9.32}{0.30} %
      & \cellstd{11.62}{0.34} %
      & \cellstd{12.19}{0.15} %
      \\
    FlexGaze-B
      & 86.0M %
      & H,K
      & \cellstd{12.47}{0.11} %
      & \cellstd{13.54}{0.16} %
      & \cellstd{12.28}{0.92} %
      & \cellstd{11.67}{0.99} %
      & \cellstd{8.71}{0.16} %
      & \cellstd{8.00}{0.44} %
      & \cellstd{12.24}{0.65} %
      & \cellstd{12.69}{0.25} %
      \\
    \bottomrule[0.8pt]
  \end{resulttabular}
  \par\vspace{3pt}
  {\footnotesize
  \begin{tabular}{@{}l@{\enspace}>{\raggedright\arraybackslash}p{0.82\textwidth}@{}}
  Preproc. & Input preparation requirements: H (head box), F (face box), L (facial landmarks), P (head pose), K (camera intrinsics). \\
  \addlinespace[2pt]
  \colorbox{gray!10}{\strut in-domain} & The model was trained on this dataset. \\
  \addlinespace[2pt]
  OmniGaze~\cite{qu2025omnigaze} & Not compared because its model weights could not be obtained and its test splits are insufficiently specified. \\
  \end{tabular}
  \par
  }
\end{table*}

\subsection{Cross-Domain Generalization}
\label{sec:experiments:generated}

To assess whether generated supervision supports cross-dataset generalization, we first evaluate the pretrained FlexGaze models directly on all eight benchmarks, without fine-tuning (Table~\ref{tab:face-detected-results}). Both models outperform the constant mean-direction baseline on seven benchmarks, indicating that the generated supervision enables transferable gaze prediction. GazeCapture is the sole exception, where neither model improves on this baseline.

Five-dataset fine-tuning has mixed effects on transfer to the excluded datasets RT-GENE, Columbia, and GFIE. Relative to their pretrained counterparts, both S and B improve on RT-GENE and Columbia but degrade on GFIE, showing that supervision from other datasets does not benefit all target domains equally. With these three datasets included in eight-dataset fine-tuning, B improves substantially over the five-dataset model on RT-GENE and GFIE. The gain on Columbia is smaller, as the five-dataset model already transfers well to this benchmark.

We further examine transfer through leave-one-dataset-out (LODO) evaluation on the five datasets, fine-tuning on four and testing on the held-out dataset in each run (Table~\ref{tab:lodo-results}). Relative to their respective pretrained versions, FlexGaze-S and FlexGaze-B achieve lower errors on GazeCapture, MPIIFaceGaze, and EYEDIAP. On Gaze360 and ETH-XGaze, which cover wider gaze angle ranges (Fig.~\ref{fig:gaze-dataset-overview}), overall errors increase for both models despite improved accuracy for gaze directed approximately toward the camera. For held-out ETH-XGaze, for example, fine-tuning lowers FlexGaze-B's error by $5.3^\circ$ for gaze within $30^\circ$ of the direction toward the camera but raises it by $6.5^\circ$ elsewhere.

\begin{table}[!tbp]
  \centering
  \footnotesize
  \setlength{\tabcolsep}{0.5pt}
  \renewcommand{\stdsuffix}[1]{\kern0.1em\raisebox{-0.4ex}{{\fontsize{6}{7}\selectfont\color{black}$\pm$#1}}}
  \DeclareRobustCommand{\lodoheader}[1]{%
    \ifresultmeasure\resultmeasurevalue{header}{\begin{tabular}[c]{@{}c@{}}#1\end{tabular}}\fi
    \begin{tabular}[c]{@{}c@{}}#1\end{tabular}}
  \renewcommand{\resultcolumnspace}{%
    \ifresultmeasure\else
      \setlength{\resultpadding}{\resultwidth{header}}%
      \ifdim\resultpadding<\resultwidth{mean}\relax
        \setlength{\resultpadding}{\resultwidth{mean}}%
      \fi
      \setlength{\resultpadding}{\dimexpr\resultwidth{mean}/2+\resultwidth{sd}-\resultpadding/2\relax}%
      \ifdim\resultpadding>0pt\hspace{\resultpadding}\fi
    \fi}
  \RenewDocumentEnvironment{resulttabular}{m +b}{%
    \stepcounter{resulttable}%
    \resultmeasuretrue
    \sbox{\resultmeasurebox}{\begin{tabular}{#1}#2\end{tabular}}%
    \resultmeasurefalse
    \sbox{\resultmeasurebox}{\begin{tabular}{#1}#2\end{tabular}}%
    \ifdim\wd\resultmeasurebox>\columnwidth
      \resizebox{\columnwidth}{!}{\usebox{\resultmeasurebox}}%
    \else
      \begin{tabular*}{\columnwidth}{#1}#2\end{tabular*}%
    \fi
  }{}
  \caption{Leave-one-dataset-out (LODO) evaluation results (mean angular error, $^\circ$).}
  \label{tab:lodo-results}
  \begin{resulttabular}{@{}l@{\hspace{3pt}}c@{\hspace{4pt}\extracolsep{\fill}}G{Gaze360}G{XGaze}G{GazeCapture}G{MPIIFaceGaze}G{EYEDIAP}@{}}
    \toprule[0.8pt]
    Method & Size & \lodoheader{Gaze360\\{\scriptsize\upshape\color{black}Face (88\%)}} & \lodoheader{XGaze}
      & \lodoheader{Gaze\\Capture} & \lodoheader{MPII\\FaceGaze} & \lodoheader{EYEDIAP} \\
    \midrule[0.8pt]
    \multicolumn{7}{l}{\textbf{Oracle baseline with constant predictions}} \\
    Mean direction & ---
      & 51.48 %
      & 42.64 %
      & \textbf{7.78} %
      & 11.75 %
      & 23.83 %
      \\
    \midrule[0.4pt]
    \multicolumn{7}{l}{\textbf{Ours, pretrained (weak labels only)}} \\
    FlexGaze & S
      & \cellstd{12.69}{0.03} %
      & \cellstd{\textbf{12.90}}{0.07} %
      & \cellstd{10.59}{0.26} %
      & \cellstd{\textbf{9.97}}{0.26} %
      & \cellstd{\textbf{8.35}}{0.27} %
      \\
    FlexGaze & B
      & \cellstd{\textbf{12.47}}{0.11} %
      & \cellstd{13.54}{0.16} %
      & \cellstd{12.28}{0.92} %
      & \cellstd{11.67}{0.99} %
      & \cellstd{8.71}{0.16} %
      \\
    \specialrule{1.0pt}{4pt}{3pt}
    \multicolumn{7}{l}{\textbf{Baseline with face-crop input}} \\
    UniGaze\pmark{*} & H & \textbf{13.12} & \textbf{11.29} & 5.13 & 5.22 & \textbf{6.14} \\
    \midrule[0.4pt]
    \multicolumn{7}{l}{\textbf{Ours, LODO fine-tuned}} \\
    FlexGaze & S
      & \cellstd{21.70}{1.45} %
      & \cellstd{16.60}{0.12} %
      & \cellstd{5.63}{0.09} %
      & \cellstd{5.57}{0.19} %
      & \cellstd{7.96}{0.13} %
      \\
    FlexGaze & B
      & \cellstd{17.74}{2.01} %
      & \cellstd{16.98}{1.62} %
      & \cellstd{\textbf{5.10}}{0.05} %
      & \cellstd{\textbf{5.14}}{0.20} %
      & \cellstd{6.83}{0.52} %
      \\
    \bottomrule[0.8pt]
  \end{resulttabular}
  \par\vspace{3pt}
  {\footnotesize
  \begin{tabular}{@{}l@{\enspace}>{\raggedright\arraybackslash}p{\dimexpr\columnwidth-2em\relax}@{}}
  $^{*}$ & UniGaze's LODO model weights are unreleased, so we quote its published results~\cite{qin2026unigaze}, obtained under a different protocol. \\
  \end{tabular}
  \par
  }
\end{table}

\subsection{Sensitivity to Crop Variations}
\label{sec:experiments:preprocessing-sensitivity}

\begin{figure}[tb]
  \centering
  \includegraphics[width=\columnwidth]{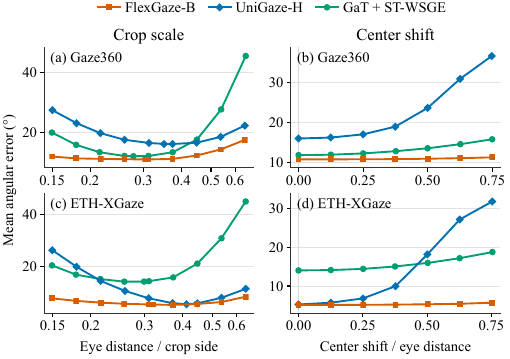}
  \caption{Sensitivity to crop scale and center shift. (a,c) Larger ratios indicate tighter crops. (b,d) Center shifts in random directions, with each method's default crop scale fixed. Eye distance is measured between the two eye centers in the image.}
  \label{fig:preprocessing-sensitivity}
\end{figure}

In practice, differences in detectors or cropping conventions can produce inputs that depart from a model's default crop. We assess robustness to these differences by independently varying crop scale and center on Gaze360 and ETH-XGaze, comparing FlexGaze-B fine-tuned on eight datasets with UniGaze-H and GaT + ST-WSGE (Fig.~\ref{fig:preprocessing-sensitivity}). Each method retains its own preprocessing pipeline. We express crop scale as the ratio of eye distance to crop side length and center shifts in units of eye distance, keeping each method's default crop scale fixed during shifts. Across both datasets, FlexGaze-B has flatter error curves overall and maintains lower mean angular errors over the tested ranges of crop scale and center shift. These results indicate that the model tolerates deviations from its default cropping settings.

\subsection{Gaze Estimation under Limited Facial Visibility}
\label{sec:experiments:limited-facial-visibility}

\begin{table}[!tbp]
  \centering
  \setlength{\tabcolsep}{2pt}
  \continuousresultbackgrounds
  \newlength{\visibilitycolorpadding}
  \setlength{\visibilitycolorpadding}{\tabcolsep}
  \def\visibilitylastcolumn{GFIE}
  \DeclareRobustCommand{\visibilityheader}[1]{\begin{tabular}[c]{@{}c@{}}#1\end{tabular}}
  \DeclareRobustCommand{\indom}[1]{\relax%
    \ifnum\pdfstrcmp{\resultcolumnname}{\visibilitylastcolumn}=0
      \resultcellbackground{\visibilitycolorpadding}{0pt}%
    \else
      \resultcellbackground{\visibilitycolorpadding}{\visibilitycolorpadding}%
    \fi
    #1}
  \RenewDocumentEnvironment{resulttabular}{m +b}{%
    \stepcounter{resulttable}%
    \resultmeasuretrue
    \sbox{\resultmeasurebox}{\begin{tabular}{#1}#2\end{tabular}}%
    \resultmeasurefalse
    \sbox{\resultmeasurebox}{\begin{tabular}{#1}#2\end{tabular}}%
    \setlength{\visibilitycolorpadding}{\dimexpr(\columnwidth-\wd\resultmeasurebox)/4\relax}%
    \ifdim\visibilitycolorpadding<0pt\setlength{\visibilitycolorpadding}{0pt}\fi
    \addtolength{\visibilitycolorpadding}{\dimexpr\tabcolsep+0.05pt\relax}%
    \begin{tabular*}{\columnwidth}{#1}#2\end{tabular*}%
  }{}
  \caption{Gaze estimation results on subsets with undetected faces (mean angular error, $^\circ$).}
  \label{tab:limited-facial-visibility-results}
  \begin{resulttabular}{@{}l@{\hspace{4pt}}c@{\hspace{4pt}}N{Params}@{\extracolsep{\fill}\hspace{4pt}}G{Gaze360}G{GFIE}@{}}
    \toprule[0.8pt]
    Method & Backbone & Params
      & \visibilityheader{Gaze360\\{\scriptsize\upshape\color{black}Face}\\{\scriptsize\upshape\color{black}undetected}}
      & \visibilityheader{GFIE\\{\scriptsize\upshape\color{black}Face}\\{\scriptsize\upshape\color{black}undetected}} \\
    \midrule[0.8pt]
    \multicolumn{5}{@{}l@{}}{\textbf{Oracle baseline with constant predictions}} \\
    Mean direction & --- & ---
      & 46.57 %
      & 42.13 %
      \\
    \midrule[0.4pt]
    \multicolumn{5}{@{}l@{}}{\textbf{Ours, pretrained (weak labels only)}} \\
    FlexGaze-S & {\scriptsize ViT-S/16} & 21.8M
      & \cellstd{22.38}{0.13} %
      & \cellstd{17.39}{0.52} %
      \\
    FlexGaze-B & {\scriptsize ViT-B/16} & 86.0M
      & \cellstd{\textbf{21.66}}{0.08} %
      & \cellstd{\textbf{15.65}}{0.03} %
      \\
    \specialrule{1.0pt}{4pt}{3pt}
    \multicolumn{5}{@{}l@{}}{\textbf{Baseline with head-crop input}} \\
    GaT + ST-WSGE & {\scriptsize Video Swin-T~\cite{liu2022videoswin}} & 27.9M
      & \indom{22.18} %
      & 21.45 %
      \\
    \midrule[0.4pt]
    \multicolumn{5}{@{}l@{}}{\textbf{Ours, fine-tuned (Gaze360 only)}} \\
    FlexGaze-S & {\scriptsize ViT-S/16} & 21.8M
      & \indom{\cellstd{21.52}{0.19}} %
      & \cellstd{19.62}{0.21} %
      \\
    FlexGaze-B & {\scriptsize ViT-B/16} & 86.0M
      & \indom{\cellstd{\textbf{20.05}}{0.37}} %
      & \cellstd{\textbf{17.01}}{0.20} %
      \\
    \specialrule{1.0pt}{4pt}{3pt}
    \multicolumn{5}{@{}l@{}}{\textbf{Ours, fine-tuned (5 sets)}} \\
    FlexGaze-S & {\scriptsize ViT-S/16} & 21.8M
      & \indom{\cellstd{21.00}{0.18}} %
      & \cellstd{18.93}{2.23} %
      \\
    FlexGaze-B & {\scriptsize ViT-B/16} & 86.0M
      & \indom{\cellstd{20.37}{0.29}} %
      & \cellstd{17.50}{1.43} %
      \\
    \midrule[0.4pt]
    \multicolumn{5}{@{}l@{}}{\textbf{Ours, fine-tuned (8 sets)}} \\
    FlexGaze-S & {\scriptsize ViT-S/16} & 21.8M
      & \indom{\cellstd{21.03}{0.25}} %
      & \indom{\cellstd{19.25}{1.50}} %
      \\
    FlexGaze-B & {\scriptsize ViT-B/16} & 86.0M
      & \indom{\cellstd{\textbf{20.33}}{0.43}} %
      & \indom{\cellstd{\textbf{15.97}}{0.27}} %
      \\
    \bottomrule[0.8pt]
  \end{resulttabular}
\end{table}

FlexGaze takes a head crop as input and can therefore estimate gaze even when the face is not detected. We evaluate this capability on the face-undetected subsets of Gaze360 and GFIE (Table~\ref{tab:limited-facial-visibility-results}). Since methods requiring face detection cannot construct inputs here, we compare with GaT + ST-WSGE, which also uses head crops. With weak supervision alone, pretrained FlexGaze-S and FlexGaze-B achieve accuracy close to GaT + ST-WSGE on Gaze360 and lower errors on GFIE. After fine-tuning on Gaze360 alone, both models outperform GaT + ST-WSGE on both subsets.

\subsection{Ablation Studies}
\label{sec:experiments:ablations}

We change one setting at a time relative to the full recipe for each training stage (Table~\ref{tab:model-ablation-results}). Fine-tuning ablations use the five-dataset setting to assess both in-domain accuracy and out-of-domain generalization.

\begin{table}[!tbp]
  \centering
  \footnotesize
  \setlength{\tabcolsep}{1pt}
  \continuousresultbackgrounds
  \DeclareRobustCommand{\ablcell}[2]{#1\stdsuffix{#2}}
  \newlength{\ablationremovedwidth}
  \newlength{\ablationremovedcandidate}
  \settowidth{\ablationremovedwidth}{\scriptsize DINOv3 init.}
  \settowidth{\ablationremovedcandidate}{\scriptsize Separate heads}
  \ifdim\ablationremovedcandidate>\ablationremovedwidth
    \setlength{\ablationremovedwidth}{\ablationremovedcandidate}
  \fi
  \settowidth{\ablationremovedcandidate}{\scriptsize FlexGaze init.}
  \ifdim\ablationremovedcandidate>\ablationremovedwidth
    \setlength{\ablationremovedwidth}{\ablationremovedcandidate}
  \fi
  \DeclareRobustCommand{\ablationreplace}[2]{$-$ \makebox[\ablationremovedwidth][l]{#1} $+$ #2}
  \caption{Pretraining and five-dataset fine-tuning ablations of FlexGaze-B (mean angular error, $^\circ$).}
  \label{tab:model-ablation-results}
  \begin{tabular*}{\columnwidth}{@{}l@{\extracolsep{\fill}}rrr@{}}
    \toprule[0.8pt]
    \multirow{2}{*}[-2.5pt]{Setting}
      & \multicolumn{3}{c}{Average over} \\
    \cmidrule(lr){2-4}
      & \multicolumn{1}{c}{5 FT sets}
      & \multicolumn{1}{c}{Other 3 sets}
      & \multicolumn{1}{c}{All 8 sets} \\
    \midrule[0.6pt]
    \multicolumn{4}{@{}l@{}}{\textbf{Pretrained (weak labels only)}} \\
    Full recipe
      & \ablcell{11.73}{0.31} %
      & \ablcell{10.97}{0.24} %
      & \ablcell{11.45}{0.11} %
      \\
    {\scriptsize $-$ Pose-unavailable samples}
      & \ablcell{\textbf{11.52}}{0.47} %
      & \ablcell{\textbf{10.80}}{0.31} %
      & \ablcell{\textbf{11.25}}{0.39} %
      \\
    {\scriptsize $-$ Crop jitter}
      & \ablcell{11.94}{0.39} %
      & \ablcell{11.61}{0.22} %
      & \ablcell{11.82}{0.18} %
      \\
    {\scriptsize \ablationreplace{DINOv3 init.}{Random init.}}
      & \ablcell{16.12}{0.21} %
      & \ablcell{16.50}{0.24} %
      & \ablcell{16.26}{0.21} %
      \\
    \midrule[0.4pt]
    \multicolumn{4}{@{}l@{}}{\textbf{Fine-tuned (5 sets shared with UniGaze)}} \\
    Full recipe
      & \indom{\ablcell{5.98}{0.07}} %
      & \ablcell{\textbf{9.69}}{0.23} %
      & \ablcell{\textbf{7.37}}{0.05} %
      \\
    {\scriptsize \ablationreplace{Separate heads}{Shared head}}
      & \indom{\ablcell{\textbf{5.86}}{0.04}} %
      & \ablcell{10.57}{0.20} %
      & \ablcell{7.63}{0.06} %
      \\
    {\scriptsize $-$ Pseudo-labeled data replay}
      & \indom{\ablcell{6.09}{0.03}} %
      & \ablcell{10.24}{0.55} %
      & \ablcell{7.64}{0.19} %
      \\
    {\scriptsize $-$ Crop jitter}
      & \indom{\ablcell{6.03}{0.03}} %
      & \ablcell{10.02}{0.06} %
      & \ablcell{7.53}{0.04} %
      \\
    {\scriptsize \ablationreplace{FlexGaze init.}{DINOv3 init.}}
      & \indom{\ablcell{6.07}{0.02}} %
      & \ablcell{10.01}{0.13} %
      & \ablcell{7.55}{0.06} %
      \\
    \bottomrule[0.8pt]
  \end{tabular*}
  \par\vspace{2pt}
  {\scriptsize\raggedright
  Columns average the per-dataset errors, measured under the protocol of Table~\ref{tab:face-detected-results}, over the five fine-tuning datasets (Gaze360, XGaze, GazeCapture, MPIIFaceGaze, EYEDIAP), the other three (Columbia, RT-GENE, GFIE), and all eight benchmarks. Each average is computed per seed and reported as mean and SD over seeds.\par}
\end{table}

\subsubsection{Pose-Unavailable Samples}
\label{sec:experiments:ablation-pose-unavailable}

The pose-unavailable branch contributes only $27\%$ of the retained pretraining labels but supplies $96\%$ of those whose gaze directions deviate by more than $150^\circ$ from the direction toward the camera. Removing this branch when pretraining FlexGaze-B slightly lowers the average errors on face-detected benchmarks (Table~\ref{tab:model-ablation-results}). On the Gaze360 and GFIE test sets, including face-undetected samples, retaining the branch reduces the Gaze360 error by $3.1^\circ$ but increases the GFIE error by $0.6^\circ$ for ground-truth gaze beyond $90^\circ$. Beyond $150^\circ$, it reduces the two errors by $10.9^\circ$ and $0.9^\circ$. The additional supervision thus mainly helps estimate gaze directed almost directly away from the camera.

\subsubsection{Crop Jitter}
\label{sec:experiments:ablation-input}

We ablate crop jitter at each training stage by disabling both crop-center sampling and scale augmentation (Table~\ref{tab:model-ablation-results}). During pretraining, crop jitter generally improves transfer to gaze benchmarks, suggesting that crop diversity supports generalization when learning from generated labels. Both fine-tuning variants start from weights pretrained with the full recipe. Continuing to use crop jitter has little effect on in-domain accuracy but improves transfer to Columbia and RT-GENE.

\subsubsection{Shared versus Separate Prediction Heads}
\label{sec:experiments:ablation-heads}

We use separate prediction heads for generated and ground-truth gaze labels to limit direct interference from pseudo-label errors. Compared with a shared head under the same initialization and replay settings, separate heads slightly raise the average error on the five fine-tuning datasets but lower it on the other three (Table~\ref{tab:model-ablation-results}), with reductions of $1.88^\circ$ on Columbia and $2.15^\circ$ on RT-GENE. This contrast suggests that better adaptation to the seen data distributions does not necessarily improve cross-domain accuracy. On GFIE, the shared head performs better and stays close to the pretrained model's error, possibly because replayed pseudo-labels still supervise it directly.

\subsubsection{Pseudo-Labeled Data Replay}
\label{sec:experiments:ablation-replay}

Without replay, five-dataset fine-tuning raises the GFIE error from the pretrained model's $12.69^\circ$ to $15.58^\circ$. Adding pseudo-labeled data replay lowers the fine-tuned model's error to $14.34^\circ$ (Table~\ref{tab:face-detected-results}). Replay therefore partly mitigates the loss of cross-domain accuracy during fine-tuning, while in-domain errors remain similar or slightly lower (Table~\ref{tab:model-ablation-results}).

\subsection{Qualitative Results}
\label{sec:experiments:qualitative}

\begin{figure}[tb]
  \centering
  \includegraphics[width=\columnwidth]{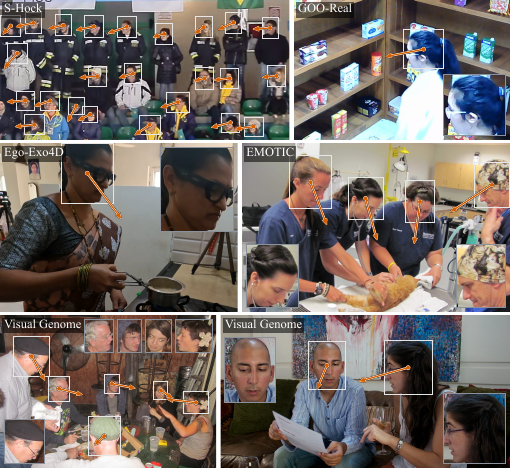}
  \caption{FlexGaze's 3D gaze predictions projected onto unseen images from S-Hock~\cite{conigliaro2015shock}, GOO-Real~\cite{tomas2021goo}, Ego-Exo4D~\cite{grauman2024egoexo4d}, EMOTIC~\cite{kosti2019emotic}, and Visual Genome~\cite{krishna2017visualgenome}. Head boxes for S-Hock and GOO-Real come from official annotations.}
  \label{fig:qualitative-results}
\end{figure}

Figure~\ref{fig:qualitative-results} presents qualitative results on datasets not used to train FlexGaze, spanning social interactions, collaborative activities, and object manipulation. The projected gaze directions point toward interaction partners or task-relevant areas. The model produces plausible predictions for distant, low-resolution heads and for people facing sideways or looking down. It also accommodates both loose and tight head boxes. These examples suggest its potential for analyzing everyday interactions and activities.

\section{Conclusion}
\label{sec:conclusion}

FlexGaze demonstrates that gaze targets and scene geometry inferred from in-the-wild images can provide effective weak supervision for 3D gaze estimation at scale, offering a way to expand training data beyond dedicated gaze collection. Pretraining on these generated labels supports cross-dataset generalization, while fine-tuning with ground-truth gaze labels yields competitive accuracy with relatively few parameters. With head crops as input, FlexGaze can estimate gaze even when no face is detected and maintains stable accuracy across a broad range of crop variations.

\section*{Acknowledgments}

Research was sponsored by the Army Research Office and was accomplished under Cooperative Agreement Number W911NF-25-2-0040. The views and conclusions contained in this document are those of the authors and should not be interpreted as representing the official policies, either expressed or implied, of the Army Research Office or the U.S. Government. The U.S. Government is authorized to reproduce and distribute reprints for Government purposes notwithstanding any copyright notation herein.

{\small
\bibliographystyle{ieee}
\bibliography{references}
}

\end{document}


\twocolumn[{\centering\LARGE\bfseries Supplementary Material\par\medskip}]
\thispagestyle{empty}

\noindent Tables, figures, sections, and equations without the prefix S refer to the main paper.

\section{Pseudo-Label Generation and Data}
\label{sec:supp:data}

This section details the label generation pipeline of Sec.~\mbox{III-A} and the pretraining corpus of Sec.~\mbox{IV-A}. The same generation and filtering recipe is applied to every corpus.

\subsection{Label Generation}
\label{sec:supp:data:generation}

We detect heads with the public YOLOv5m~\cite{jocher2021yolov5} weights trained on CrowdHuman~\cite{shao2018crowdhuman}\footnote{Weights released at \href{https://github.com/MahenderAutonomo/yolov5-crowdhuman}{github.com/MahenderAutonomo/yolov5-crowdhuman}.} at three input resolutions. The detections at each resolution are filtered by confidence and non-maximum suppression (NMS), mapped back to the original image, and merged across resolutions, after which boxes nested inside a larger box are removed. Table~\ref{tab:supp-label-settings} gives the values.

For each head box, PaGE~\cite{ye2026page} takes the scene image and the head box and predicts a gaze target heatmap. We apply a softmax with temperature $0.06$ to the heatmap logits. We then average the centers of the cells in the $3\times3$ neighborhood of the most probable cell, weighted by their probabilities, using only cells inside the heatmap. Scaling this point to the image size gives the 2D target.

UniDepthV2~\cite{piccinelli2026unidepthv2,piccinelli2024unidepth} with the public ViT-L/14 weights predicts a metric depth map and camera intrinsics $K$ from the full image, with depth measured along the optical axis. At the head box center and at the 2D target, we take the median of the valid depths in a $3\times3$ window, denoted $z_h$ and $z_t$. With homogeneous pixel coordinates $\tilde{\mathbf{u}}_h$ and $\tilde{\mathbf{u}}_t$, the gaze origin and target are
\begin{equation}
  \mathbf{o}=z_hK^{-1}\tilde{\mathbf{u}}_h,\qquad
  \mathbf{t}=e^{-0.03}z_tK^{-1}\tilde{\mathbf{u}}_t,
  \label{eq:supp:backprojection}
\end{equation}
where $e^{-0.03}$ is the target depth correction described below, and the gaze label is $\mathbf{g}=(\mathbf{t}-\mathbf{o})/\|\mathbf{t}-\mathbf{o}\|_2$. The $x$, $y$, and $z$ camera axes point right, down, and forward, respectively.

Head pose comes from the 3DDFA-V3 pipeline~\cite{wang2024geometric,lei2023hrn,deng2020retinaface}, whose own face detection and five-point alignment locate faces independently of our head boxes. We therefore match the fits to the head boxes. With $d$ the distance between the centers of a fitted face window and a head box of width $w$ and height $h$, the center agreement is $q=(1+d/\sqrt{wh})^{-1}$. We solve a one-to-one assignment within each image using the center agreement $q$, subject to a lower bound on $q$. A match whose fitted window is centered outside the head box is rejected if that center lies inside another head box or if only a small part of the window lies inside the matched head box. Matched heads use the fitted head rotation, and their face center is the mean 3D position of the four eye corners and two mouth corners. All other heads, including those with rejected matches, enter the pose-unavailable branch.

\subsection{Target Depth Correction and Filtering}
\label{sec:supp:data:filtering}

\begin{table}[t]
  \centering
  \footnotesize
  \setlength{\tabcolsep}{0pt}
  \newlength{\supplabellabelwidth}
  \settowidth{\supplabellabelwidth}{Pose-unavailable branch}
  \caption{Settings for label generation, target depth correction, and filtering.}
  \label{tab:supp-label-settings}
  \begin{tabular}{@{}>{\raggedright\arraybackslash}p{\supplabellabelwidth}@{\hspace{6pt}}>{\raggedright\arraybackslash}p{\dimexpr\columnwidth-\supplabellabelwidth-6pt\relax}@{}}
    \toprule
    Item & Setting \\
    \midrule
    \multicolumn{2}{@{}l}{\textit{Head detection}} \\
    Detector & YOLOv5m trained on CrowdHuman \\
    Input resolutions & longest side of $256$, $320$, and $896$ px \\
    Per resolution & confidence $\geq0.05$, NMS at IoU $0.45$, at most $300$ boxes \\
    Across resolutions & confidence $\geq0.7$, NMS at IoU $0.45$ \\
    Nested boxes & removed if $\geq80\%$ of the box area lies inside a larger box \\
    \midrule
    \multicolumn{2}{@{}l}{\textit{Gaze target and scene geometry}} \\
    Target readout & PaGE heatmap, softmax at temperature $0.06$, probability-weighted centroid of the $3\times3$ cells around the peak \\
    Depth and intrinsics & UniDepthV2 ViT-L/14, median of valid depths in a $3\times3$ window \\
    Target depth correction & target depth scaled by $e^{-0.03}$ \\
    \midrule
    \multicolumn{2}{@{}l}{\textit{Basic checks}} \\
    Head box size & shorter side $\geq16$ px \\
    PaGE confidence & in-frame probability $\geq0.5$, heatmap peak probability $\geq0.2$ \\
    Raw depth & valid and $\geq0.1$ m at head box center and target \\
    Physical head box size & larger side from $35$ to $1600$ mm \\
    Head-to-target distance & $\geq0.05$ m after correction \\
    \midrule
    \multicolumn{2}{@{}l}{\textit{Head pose matching}} \\
    Center agreement & one-to-one assignment with $q\geq0.5$ \\
    Fit centered outside head box & rejected if its center lies inside another head box or less than $25\%$ of the fit window lies inside the head box \\
    \midrule
    \multicolumn{2}{@{}l}{\textit{Direction filters}} \\
    Pose-available branch & $(a/60^\circ)^2+(b/V)^2\leq1$ with $V=60^\circ$ upward and $V=80^\circ$ downward \\
    Pose-unavailable branch & angle to the head-to-camera direction $>30^\circ$ \\
    \bottomrule
  \end{tabular}
\end{table}

Table~\ref{tab:supp-label-settings} summarizes all generation and filtering settings. The target depth correction scales the target depth along its camera ray by the fixed factor $e^{-0.03}$, which leaves the 2D target and the 3D gaze origin unchanged. The factor was chosen based on results on the GFIE~\cite{hu2023gfie} validation split. On the GFIE training split, which is not used to select the factor, correction lowers the mean gaze error from $22.13^\circ$ to $18.70^\circ$ on the same $58{,}957$ samples before filtering. Restricted to the $47{,}534$ samples that pass filtering both with and without correction, a subset of the filtered labels in Table~I, it lowers the error from $18.57^\circ$ to $15.25^\circ$. The depth validity and physical size checks use the raw depths, whereas the head-to-target distance check, the direction filters, and training use the corrected labels.

The basic checks exclude samples with insufficient evidence or clearly implausible geometry. The physical head box size is estimated as $s=z_h\max(w/f_x,h/f_y)$, where $f_x$ and $f_y$ are the focal lengths and $w$ and $h$ the head box width and height, all in pixels. The accepted range leaves a wide margin for variation in head box extent, subjects, and pose, so that the check targets grossly wrong scale estimates.

When head pose is available, we rotate the corrected gaze direction into the head frame and compute its total deviation $\theta$ from the head's forward direction. We decompose $\theta$ along the deviation direction into a horizontal component $a$ and a vertical component $b$ with $a^2+b^2=\theta^2$, which differ from separately computed yaw and pitch angles. A sample is kept if $(a/60^\circ)^2+(b/V)^2\leq1$, with $V=60^\circ$ for upward and $V=80^\circ$ for downward deviations. The asymmetric vertical bound reflects the smaller upward than downward range of eye rotation~\cite{lee2019eyemovement,gerling1997duction}, with allowance for pose estimation errors~\cite{martyniuk2022dad3dheads}. These bounds serve as a coarse filter rather than precise physiological limits.

When head pose is unavailable, the generated gaze must deviate by more than $30^\circ$ from the direction from the head box center toward the camera. This condition excludes heads whose gaze points roughly toward the camera, because such views usually show a face that the fitting should recover, and retains rear views, extreme head poses, and facial occlusion.

The sample weights of pseudo-labels use PaGE's two confidences. Let $p_i^{\mathrm{in}}$ be the in-frame probability and $p_i^{\mathrm{peak}}=\sigma(\max_{\mathbf{u}}h_i(\mathbf{u}))$ the heatmap peak probability, where $h_i$ denotes the raw heatmap logits and $\sigma$ the sigmoid function. This peak probability is independent of the tempered softmax used for target readout, and the $0.2$ floor in the basic checks applies to it. The raw weight is
\begin{equation}
  r_i=\max(p_i^{\mathrm{in}},10^{-3})\,\max(p_i^{\mathrm{peak}},10^{-3}).
  \label{eq:supp:raw-weight}
\end{equation}
During pretraining, the weights are normalized within each global batch $\mathcal{B}$ as $w_i=|\mathcal{B}|\,r_i/\sum_{j\in\mathcal{B}}r_j$, which gives a mean of one. During fine-tuning, the same normalization is applied within the pseudo-labeled part of each batch. Depth confidence does not enter the weights.

\subsection{Pretraining Corpus and Deduplication}
\label{sec:supp:data:corpus}

The pretraining corpus consists of the images of the $17$ public datasets~\cite{kuznetsova2020openimages,rothe2018imdbwiki,rothe2015dex,recasens2015where,kuprashevich2024mivolo,kuprashevich2025beyond,zhang2018expw,yang2016widerface,dhall2018emotiw,young2014flickr30k,chong2020attended,andriluka2014mpiihumanpose,karras2019stylegan,karkkainen2021fairface,liu2015celeba,yang2022heco,kumar2020luvli,koestinger2011aflw,liu2020lapa,deng2019menpo,zafeiriou2017menpo} listed in Sec.~\mbox{IV-A}, without their original task annotations. GazeFollow and VAT are restricted to their training splits. For WIDER FACE and GAF 3.0, we include the released training, validation, and test images. We use the in-the-wild images of FFHQ and the unaligned original images of CelebA. Frames from the same video in VAT and MPII Human Pose are used as separate images without further frame sampling, and every epoch reads the same set of samples.

We deduplicate whole images within and across all $17$ corpora. Deduplication runs after all basic checks other than the heatmap peak floor. The heatmap peak floor and the direction filters are then applied to the deduplicated samples. Two images are duplicates if the SHA-256 digests of their decoded pixels match or if their $64$-bit perceptual hashes (pHash)~\cite{zauner2011rihamark}, computed from discrete cosine transform coefficients, are identical. Duplicate relations are merged into connected groups, and each group keeps only the copy with the most pixels. All head instances in the discarded copies are removed.

Table~\ref{tab:supp-corpus-retention} reports the retained counts for each corpus at each stage of label generation and filtering. Deduplication removes the largest shares from IMDB-WIKI, VAT, MERL-RAV, and HECO. Of the $3{,}242{,}122$ head instances used for pretraining, $2{,}371{,}858$ are retained through the pose-available branch and $870{,}264$ ($27\%$) through the pose-unavailable branch.

\begin{table}[t]
  \centering
  \footnotesize
  \setlength{\tabcolsep}{2pt}
  \caption{Head instance retention per pretraining corpus.}
  \label{tab:supp-corpus-retention}
  \begin{tabular*}{\columnwidth}{@{}l@{\extracolsep{\fill}}rrrrrr@{}}
    \toprule
    Corpus & \shortstack{Images\\(k)} & \multicolumn{4}{c}{Head instances (k)} & \shortstack{Keep\\(\%)} \\
    \cmidrule(lr){3-6}
      & & Raw & Valid & Dedup. & Used & \\
    \midrule
    Open Images
      & 1,910.1 & 4,565.9
      & 2,705.7
      & 2,670.4
      & 2,071.0
      & 45.4
      \\
    IMDB-WIKI
      & 523.1 & 1,560.2
      & 744.5
      & 408.4
      & 334.0
      & 21.4
      \\
    GazeFollow (train)
      & 119.1 & 436.3
      & 360.0
      & 354.3
      & 275.6
      & 63.2
      \\
    LAGENDA
      & 67.2 & 331.4
      & 174.2
      & 172.2
      & 140.8
      & 42.5
      \\
    ExpW
      & 107.0 & 303.0
      & 131.8
      & 125.7
      & 99.8
      & 32.9
      \\
    WIDER FACE
      & 32.2 & 215.9
      & 112.4
      & 112.1
      & 85.4
      & 39.6
      \\
    GAF 3.0
      & 17.2 & 164.4
      & 60.6
      & 59.4
      & 46.2
      & 28.1
      \\
    Flickr30k
      & 31.8 & 117.6
      & 56.3
      & 55.6
      & 41.1
      & 35.0
      \\
    VAT (train)$^\dagger$
      & 58.5 & 221.6
      & 140.2
      & 45.5
      & 37.8
      & 17.0
      \\
    MPII Human Pose$^\dagger$
      & 25.0 & 63.3
      & 49.7
      & 49.5
      & 37.5
      & 59.2
      \\
    FFHQ
      & 70.0 & 150.0
      & 45.7
      & 43.6
      & 31.2
      & 20.8
      \\
    FairFace
      & 97.7 & 155.9
      & 33.0
      & 32.9
      & 17.8
      & 11.4
      \\
    CelebA
      & 202.6 & 221.8
      & 15.9
      & 15.7
      & 10.2
      & 4.6
      \\
    HECO
      & 5.5 & 25.2
      & 13.8
      & 7.0
      & 5.8
      & 22.9
      \\
    MERL-RAV
      & 19.3 & 46.2
      & 13.3
      & 6.4
      & 5.0
      & 10.8
      \\
    LaPa
      & 22.2 & 30.0
      & 3.7
      & 3.6
      & 1.9
      & 6.4
      \\
    Menpo2D
      & 16.2 & 19.1
      & 1.8
      & 1.8
      & 1.0
      & 5.4
      \\
    \midrule
    Total
      & 3,324.6 & 8,627.6
      & 4,662.6
      & 4,164.1
      & 3,242.1
      & 37.6
      \\
    \bottomrule
  \end{tabular*}
  \par\vspace{3pt}
  \begin{minipage}{\columnwidth}
    \footnotesize
    Rows are sorted by Used. Images counts the images from the selected splits before deduplication. The other columns count head instances. Raw counts detected heads, Valid those passing the basic checks other than the heatmap peak floor, Dedup.\ those remaining after image-level deduplication, and Used those further passing the heatmap peak floor and the direction filters. Keep is Used/Raw. VAT denotes VideoAttentionTarget, and $^\dagger$ marks video frames. A single video may contribute multiple frames.
  \end{minipage}
\end{table}

\section{Training Details}
\label{sec:supp:training}

This section details the input construction, model, and learning procedure of Secs.~\mbox{III-B} and~\mbox{III-C} and the training setup of Sec.~\mbox{IV-A}. Table~\ref{tab:supp-training-settings} summarizes the settings of both training stages.

\begin{table}[t]
  \centering
  \footnotesize
  \setlength{\tabcolsep}{0pt}
  \newlength{\supptrainlabelwidth}
  \newlength{\supptrainfinewidth}
  \newlength{\supptrainprewidth}
  \settowidth{\supptrainlabelwidth}{Resolution degradation}
  \settowidth{\supptrainfinewidth}{two, ground truth and replay}
  \setlength{\supptrainprewidth}{\dimexpr\columnwidth-\supptrainlabelwidth-\supptrainfinewidth-12pt\relax}
  \newcommand{\both}[1]{\multicolumn{2}{>{\raggedright\arraybackslash}p{\dimexpr\supptrainprewidth+\supptrainfinewidth+6pt\relax}@{}}{#1}}
  \caption{Training settings for pretraining and fine-tuning.}
  \label{tab:supp-training-settings}
  \begin{tabular}{@{}p{\supptrainlabelwidth}@{\hspace{6pt}}>{\raggedright\arraybackslash}p{\supptrainprewidth}@{\hspace{6pt}}>{\raggedright\arraybackslash}p{\supptrainfinewidth}@{}}
    \toprule
    Setting & Pretraining & Fine-tuning \\
    \midrule
    Backbone & \both{ViT-S/16 (S) or ViT-B/16 (B), $224\times224$ input} \\
    Inference parameters & \both{$21.8$M (S), $86.0$M (B)} \\
    Initialization & DINOv3 & averaged pretrained weights \\
    Prediction heads & one & two, ground truth and replay \\
    Optimizer & \both{AdamW, weight decay $0.05$} \\
    Learning rate & \both{heads $2\times10^{-4}$, backbone $2\times10^{-5}$, constant} \\
    Global batch size & \both{$4096$} \\
    Precision & \both{BF16 mixed precision} \\
    Epochs & $30$ (S), $20$ (B) & $10$ \\
    Averaged epochs & $16$--$30$~(S), $11$--$20$~(B) & $6$--$10$ \\
    Replay share & --- & $50\%$ of the samples per epoch \\
    Weight normalization & whole batch & pseudo-labeled part \\
    Seeds & \both{$0$, $1$, $2$} \\
    \midrule
    \multicolumn{3}{@{}l}{\textit{Crop jitter and camera roll}} \\
    Camera roll & \both{uniform in $[-15^\circ,15^\circ]$} \\
    Center sampling & \both{ellipse extending $0.1\ell$ beyond each focus} \\
    Crop scale & \both{$[0.9,1.1]$ at head box center, $[0.5,1.5]$ at face center} \\
    Minimum crop side & \both{$\min(16\,\mathrm{px},w_0)$, with $w_0$ the default crop side in the original image} \\
    \midrule
    \multicolumn{3}{@{}l}{\textit{Appearance augmentation}} \\
    Horizontal flip & \both{probability $0.5$} \\
    Color jitter & \both{probability $0.8$, factors in $[0.7,1.3]$, hue in $[-9^\circ,9^\circ]$} \\
    Gaussian blur & \both{probability $0.3$, $\sigma\in[0.001,2]$ px} \\
    Resolution degradation & \both{probability $0.3$, scale in $[0.35,1)$} \\
    \bottomrule
  \end{tabular}
  \vspace{2pt}
  \begin{minipage}{\columnwidth}
    \footnotesize
    $\ell$ denotes the shorter side of the head box projected onto the virtual image plane. Color jitter factors apply to brightness, contrast, and saturation. Crop jitter and camera roll are not applied to inputs read directly from released crops or face patches.
  \end{minipage}
\end{table}

\subsection{Model and Optimization}
\label{sec:supp:training:model}

FlexGaze-S and FlexGaze-B use ViT-S/16 and ViT-B/16 backbones initialized with the public DINOv3 weights~\cite{simeoni2025dinov3}, and the backbones are not frozen. The prediction head takes the class token feature and applies LayerNorm, a linear layer with $512$ outputs, GELU, a second LayerNorm, and a linear layer with $4$ outputs. The first three outputs are $\ell_2$-normalized to give the direction $\hat{\mathbf{g}}$, and the fourth passes through a softplus with an offset of $0.01$ to give the concentration $\kappa$. We evaluate the vMF negative log-likelihood of Eq.~(1) in the numerically stable equivalent form
\begin{equation}
  \begin{aligned}
    \ell_i={}&-\log\kappa_i+\log(2\pi)+\log(1-e^{-2\kappa_i})\\
    &+\kappa_i(1-\hat{\mathbf{g}}_i^\top\mathbf{g}_i).
  \end{aligned}
  \label{eq:supp:vmf}
\end{equation}

Both stages use AdamW~\cite{loshchilov2019adamw} with a weight decay of $0.05$ on all parameters. The learning rates are $2\times10^{-4}$ for the heads and $2\times10^{-5}$ for the backbone, constant throughout training without warmup or decay. Training uses BF16 mixed precision and a global batch size of $4096$.

We train independently with seeds $0$, $1$, and $2$. Within each seed, we average with equal weights the checkpoints from the second half of each stage, namely epochs $16$ to $30$ for FlexGaze-S and $11$ to $20$ for FlexGaze-B in pretraining, and epochs $6$ to $10$ in fine-tuning. These windows are fixed in advance, and no checkpoint is selected on validation results. The averaged pretrained weights produce the reported pretrained results and initialize fine-tuning for the same size and seed.

Training runs on NVIDIA H100 80 GB GPUs. Each job uses $2$ to $16$ GPUs depending on availability, and the global batch size stays at $4096$. One pretraining epoch typically takes about $8$ minutes on eight GPUs for either model size, and the $30$ pretraining epochs of FlexGaze-S take about $4$ hours.

\subsection{Input Construction and Augmentation}
\label{sec:supp:training:input}

Given the homogeneous pixel coordinates $\tilde{\mathbf{u}}$ of the crop center, the virtual camera's forward axis is the unit ray $\mathbf{f}=K^{-1}\tilde{\mathbf{u}}/\|K^{-1}\tilde{\mathbf{u}}\|_2$. Using the original camera's horizontal axis $\mathbf{e}_x=(1,0,0)^\top$ as reference, the downward axis is $\mathbf{d}=(\mathbf{f}\times\mathbf{e}_x)/\|\mathbf{f}\times\mathbf{e}_x\|_2$ and the rightward axis is $\mathbf{r}=\mathbf{d}\times\mathbf{f}$, giving the rotation $R=(\mathbf{r},\mathbf{d},\mathbf{f})^\top$. We rotate the rays through the midpoints of the four head box edges by $R$ and project them onto the virtual camera's unit-depth plane. The smaller of the projected width and height defines the side $\ell$ of a square window, which is resampled bilinearly from the original image to a $224\times224$ input. Labels are expressed in the virtual camera frame as $R\mathbf{g}$, and predictions are mapped back with $R^\top$ at inference. Constructing the input requires neither head pose nor depth.

Crop jitter samples the crop center on the unit-depth plane of the virtual camera facing the original head box center. Let $\mathbf{q}_h$ and $\mathbf{q}_f$ be the projections of the head box center and the face center onto this plane, $r=0.1\ell$, and $c=\|\mathbf{q}_f-\mathbf{q}_h\|_2/2$. The sampling region is the ellipse with foci $\mathbf{q}_h$ and $\mathbf{q}_f$, semi-major axis $c+r$, and semi-minor axis $\sqrt{r^2+2cr}$, so that its major axis extends $r$ beyond each focus. We sample uniformly over its area. Without a face center, the region reduces to a disk of radius $r$ around $\mathbf{q}_h$. The region depends only on the original head box and face center, not on the subsequently sampled scale. In pretraining, the face center comes from the matched 3DDFA-V3 fits of Sec.~\ref{sec:supp:data:generation}.

For a sampled center $\mathbf{q}$, its relative position along the line between the two centers is
\begin{equation}
  t=\min\!\left(1,\max\!\left(0,\frac{(\mathbf{q}-\mathbf{q}_h)^\top(\mathbf{q}_f-\mathbf{q}_h)}{\|\mathbf{q}_f-\mathbf{q}_h\|_2^2}\right)\right),
  \label{eq:supp:scale-position}
\end{equation}
with $t=0$ when the face center is unavailable or coincides with the head box center. The scale is drawn uniformly from $[0.9-0.4t,\,1.1+0.4t]$, which interpolates linearly between $[0.9,1.1]$ at the head box center and $[0.5,1.5]$ at the face center, and the same scale applies to the window width and height. With $w_0$ the default window side in pixels of the original image, the lower scale bound is first raised to at least $\min(16,w_0)/w_0$. Downscaling thus never yields a window below $16$ pixels, and windows already below $16$ pixels are not shrunk further. We set these ranges manually after inspecting example crops.

The camera roll $\alpha$ is drawn uniformly from $[-15^\circ,15^\circ]$. With $R_{\mathbf{q}}$ the normalization rotation toward the sampled center, the augmented view is resampled from the original image with the composed rotation $R_z(\alpha)R_{\mathbf{q}}$, and the label becomes $R_z(\alpha)R_{\mathbf{q}}\mathbf{g}$. The 3D gaze origin remains at the original head box center regardless of the crop center. Table~\ref{tab:supp-training-settings} lists the appearance augmentations. Horizontal flipping also negates the horizontal component of the label, whereas the other appearance augmentations leave the label unchanged. Brightness, contrast, and saturation factors are drawn independently, the blur standard deviation refers to the $224\times224$ input, and resolution degradation downsamples the input and resizes it back. Both stages use the same augmentation parameters, with the fine-tuning exceptions described below.

\subsection{Fine-Tuning with Pseudo-Labeled Data Replay}
\label{sec:supp:training:finetuning}

Fine-tuning starts from the averaged pretrained weights for the same size and seed. Both prediction heads have the pretrained head's architecture and are initialized with its weights. The heads are then updated separately, and both losses update the shared backbone. Inference uses only the backbone and the head supervised by ground-truth labels.

Each fine-tuning epoch makes one pass over the training data with ground-truth labels, with EYEDIAP~\cite{funesmora2014eyediap} frames sampled as described below. It also draws an equal number of replay samples from the $3{,}242{,}122$ pseudo-labeled pretraining samples. Replay therefore accounts for $50\%$ of the samples processed in each epoch. Replayed samples keep the pretraining labels, filtering, and augmentation. Each corpus receives a share of the replay samples proportional to its size and is read in shuffled passes, without repeating a sample before its pass is complete. The $50\%$ share holds per epoch, while the fraction of pseudo-labeled samples in each batch fluctuates around one half. Fine-tuning minimizes Eq.~(2) without an additional weight on the replay loss.

Fine-tuning inputs depend on the data format. For ETH-XGaze~\cite{zhang2020ethxgaze} and Gaze360~\cite{kellnhofer2019gaze360}, we use both the released crops and inputs that we construct from the original images with detected head boxes. For RT-GENE~\cite{fischer2018rtgene}, we use the released narrow and wide face patches, each in its original and inpainted versions. GazeCapture~\cite{krafka2016eyetracking}, MPIIFaceGaze~\cite{zhang2017fullface}, EYEDIAP, and Columbia~\cite{smith2013gazelocking} use head boxes that we detect in the original images, and GFIE uses its released head boxes. For EYEDIAP, we use the continuous screen target (CS) and floating target (FT) sessions and read every $5$th and every $3$rd frame of each recording per epoch, respectively. The starting offset advances by one frame per epoch, so consecutive epochs cover all frames. All frames of the other video datasets are read in every epoch.

Inputs constructed from the original images with head boxes use the pretraining settings of crop jitter and camera roll. For ETH-XGaze and MPIIFaceGaze, the face center is the image-plane mean of the released 2D landmarks of the four eye corners and two mouth corners. For the other datasets, it is the projected mean 3D position of the same six points from 3DDFA-V3 fits, matched to the head boxes with the checks used in pretraining. Without a face center, sampling uses the disk around the head box center and the scale range at the head box center. Released crops and face patches keep their original framing without crop jitter or camera roll but receive all appearance augmentations. Face centers are used only for crop jitter during training and are not needed at inference.

\section{Evaluation Protocols}
\label{sec:supp:evaluation}

This section details the comparison protocol of Sec.~\mbox{IV-C}. We define the evaluation subsets and report their sample counts, summarize the settings of each dataset, and describe how we reproduce the compared methods. We then define the constant mean-direction baseline and specify the settings of the crop sensitivity experiment in Sec.~\mbox{IV-F}.

\subsection{Evaluation Subsets}
\label{sec:supp:evaluation:subsets}

Every comparison reports the mean angular error over common samples, which are the test samples for which every participating model completes input preprocessing and gaze prediction. Gaze360 and GFIE contain rear views and other samples without a visible face, so we divide their test samples according to face detection and landmark localization. A sample becomes a face-detected candidate only if it passes three checks on the annotated subject. First, SCRFD~\cite{guo2021scrfd} (det\_10g, confidence threshold $0.5$, NMS threshold $0.4$) must return a face box containing the annotated midpoint between the subject's eyes. Detection uses the released head crop for Gaze360 and the full image for GFIE. Second, the RetinaFace~\cite{deng2020retinaface} detector of the official L2CS-Net~\cite{abdelrahman2023l2cs} pipeline must also return such a box. Third, FAN~\cite{bulat2017facealignment} landmarks on the subject's face must yield a non-empty eye crop under the eye-cropping rule of 3DGazeNet~\cite{ververas2024gazenet}. The face-detected subset consists of the candidates for which every participating model produces a prediction and is used in Table~II. Test samples that fail any of the three checks are face-undetected candidates, and those for which both GaT + ST-WSGE~\cite{vuillecard2025gat} and FlexGaze produce predictions form the face-undetected subset used in Table~IV. Detection only determines which samples are counted and does not change the input of any method.

The other six datasets were recorded with the face in view, so we apply no face-detection check and all of their test samples enter the computation of common samples. On these datasets, the FAN check removes only three RT-GENE samples. A method produces no prediction when its own pipeline cannot construct an input. This happens when its detector finds no face box or head box containing the subject. It also happens when the crop window of FlexGaze extends beyond the released body crop of Gaze360 or when the released annotation of the original ETH-XGaze images gives no head placement for a frame. Methods with normalized face inputs produce predictions for every candidate sample. The common samples also require predictions from the further released weights that we evaluate under the same protocol (Table~\ref{tab:supp-full-comparison}), which removes two more ETH-XGaze samples than using only the models in Table~II. Tables~\ref{tab:supp:face-detected-counts} and~\ref{tab:supp:face-undetected-counts} report the sample counts at each step.

\begin{table*}[!tbp]
  \centering
  \footnotesize
  \setlength{\tabcolsep}{3pt}
  \caption{Sample counts for the comparison on face-detected samples (Table~II).}
  \label{tab:supp:face-detected-counts}
  \begin{tabular*}{\textwidth}{@{\extracolsep{\fill}}lrrrrrrrrrrr@{}}
    \toprule
    & & & & \multicolumn{6}{c}{Samples without a prediction} & & \\
    \cmidrule(lr){5-10}
    Test set & Total & SCRFD & FAN
      & \begin{tabular}[c]{@{}c@{}}L2CS-Net\\(RetinaFace)\end{tabular}
      & \begin{tabular}[c]{@{}c@{}}GaT +\\ST-WSGE\end{tabular}
      & \begin{tabular}[c]{@{}c@{}}Norm.\\face\end{tabular}
      & FlexGaze
      & 3DGazeNet
      & GeomGaze
      & Common
      & \begin{tabular}[c]{@{}c@{}}Retained\\(\%)\end{tabular} \\
    \midrule
    Gaze360      & 25,969  & 23,181 & 9  & 226 & 6     & 0 & 123   & 0   & 25    & 22,831  & 87.9 \\
    ETH-XGaze    & 159,786 & ---    & 0  & 828 & 2,723 & 0 & 804   & 185 & 804   & 156,697 & 98.1 \\
    GazeCapture  & 164,209 & ---    & 0  & 2   & 3,952 & 0 & 248   & 1   & 248   & 160,134 & 97.5 \\
    MPIIFaceGaze & 9,055   & ---    & 0  & 0   & 33    & 0 & 181   & 0   & 181   & 8,867   & 97.9 \\
    EYEDIAP      & 7,871   & ---    & 0  & 0   & 0     & 0 & 0     & 0   & 0     & 7,871   & 100.0 \\
    Columbia     & 1,995   & ---    & 0  & 0   & 0     & 0 & 0     & 0   & 0     & 1,995   & 100.0 \\
    RT-GENE      & 38,850  & ---    & 3  & 0   & 1,920 & 0 & 1,919 & 0   & 1,919 & 36,927  & 95.1 \\
    GFIE         & 6,281   & 5,441  & 10 & 24  & 0     & 0 & 0     & 0   & 0     & 5,407   & 86.1 \\
    \bottomrule
  \end{tabular*}
  \par\vspace{3pt}
  \begin{minipage}{\textwidth}
    Total is the number of test samples, counting for GazeCapture the frames covered by the FAZE geometry (out of 179,496). SCRFD is the number of samples in which SCRFD detects the subject's face, checked on Gaze360 and GFIE only. FAN is the number of samples removed because FAN landmarks yield no valid (non-empty) eye crop. Each method column counts the samples passing these checks for which the method produces no prediction, and columns may overlap. The GaT + ST-WSGE column also covers the GaT baseline, which reads the same inputs. Norm.\ face denotes PureGaze, the XGaze baseline, GazeTR-Hybrid, and all UniGaze weights. 3DGazeNet and GeomGaze appear in Table~\ref{tab:supp-full-comparison}. Common is the number of samples with predictions from every method, and Retained is Common as a percentage of Total.
  \end{minipage}
\end{table*}

\begin{table}[!tbp]
  \centering
  \footnotesize
  \setlength{\tabcolsep}{3pt}
  \caption{Sample counts for the comparison on face-undetected samples (Table~IV).}
  \label{tab:supp:face-undetected-counts}
  \begin{tabular*}{\columnwidth}{@{\extracolsep{\fill}}lrrrrrr@{}}
    \toprule
    & & & \multicolumn{2}{c}{No prediction} & & \\
    \cmidrule(lr){4-5}
    Test set & Total
      & \begin{tabular}[c]{@{}c@{}}Face\\undetected\end{tabular}
      & \begin{tabular}[c]{@{}c@{}}GaT +\\ST-WSGE\end{tabular}
      & FlexGaze
      & Common
      & \begin{tabular}[c]{@{}c@{}}Retained\\(\%)\end{tabular} \\
    \midrule
    Gaze360 & 25,969 & 3,023 & 85 & 124 & 2,876 & 11.1 \\
    GFIE    & 6,281  & 874   & 0  & 0   & 874   & 13.9 \\
    \bottomrule
  \end{tabular*}
  \par\vspace{3pt}
  \begin{minipage}{\columnwidth}
    Face undetected counts the test samples outside the face-detected candidates (Table~\ref{tab:supp:face-detected-counts}). Common counts the samples with predictions from both methods, and Retained gives Common as a percentage of Total.
  \end{minipage}
\end{table}

\subsection{Dataset-Specific Protocols}
\label{sec:supp:evaluation:datasets}

Each method predicts a unit gaze direction $\hat{\mathbf{g}}_m$ from its own origin. This origin is the head box center for FlexGaze, the face center for normalized-face methods, and the midpoint between the eyes for GaT + ST-WSGE and L2CS-Net. On datasets with 3D gaze targets, we place the method origin $\mathbf{o}_m$ on the camera ray through its image location, at the depth of the dataset origin $\mathbf{o}_d$. With the ground-truth target $\mathbf{t}$, the converted prediction is
\begin{equation}
  \hat{\mathbf{g}}_{m\rightarrow d}=\frac{\hat{\mathbf{p}}_m-\mathbf{o}_d}{\|\hat{\mathbf{p}}_m-\mathbf{o}_d\|},
  \qquad
  \hat{\mathbf{p}}_m=\mathbf{o}_m+\|\mathbf{t}-\mathbf{o}_m\|\,\hat{\mathbf{g}}_m,
  \label{eq:supp:origin-conversion}
\end{equation}

\noindent That is, we take the point on the predicted ray whose distance from the method origin equals the ground-truth target distance. We then compare the direction from the dataset origin to this point with the ground-truth direction from $\mathbf{o}_d$ to $\mathbf{t}$. The conversion leaves a prediction unchanged when the two origins coincide. The ground-truth origin depth and target distance enter only this conversion after prediction, which keeps depth estimation errors out of the comparison. All directions are expressed in the original camera frame. Below we list the split and frame selection of each dataset, and every frame of a test split is evaluated unless stated otherwise. Our evaluation code identifies the test samples selected by these rules. The distributions in Figs.~3 and~4 include all test samples listed below, before the face-detection checks and common-sample selection, except that EYEDIAP is shown at the full frame rate over all $118{,}295$ valid frames.

\subsubsection{Gaze360}

We use the official test split with $25{,}969$ samples, and the dataset origin is the released midpoint between the eyes. Gaze360 releases neither camera intrinsics nor head pose, so we recover the pinhole camera of each sensor from the released 3D eye positions and their image locations. Methods that need head pose or a face center obtain them by fitting the ETH-XGaze face model with PnP~\cite{lepetit2009epnp} to landmarks estimated within the SCRFD face box. Methods that need head boxes detect them in the released body crops.

\subsubsection{ETH-XGaze}

We evaluate on all $159{,}786$ frames of the $15$ test subjects and compute errors locally from the released test annotations. The dataset origin is the face center of the official normalization. Methods trained on the official normalized images read the official $224$-pixel normalized patches, except 3DGazeNet, which reads the official $448$-pixel normalized faces it was trained on, and methods that read raw images use the released camera calibration. For fine-tuning on raw ETH-XGaze images, we use the revised annotations and camera calibration from the authors' XGaze3D release~\cite{qin2025xgaze3d}\footnote{\url{https://github.com/ut-vision/XGaze3D}}. The branch using official normalized images retains the original release data. The XGaze3D files we use do not cover the test subjects, so evaluation uses the original ETH-XGaze annotations and calibration.

\subsubsection{GazeCapture}

We use the official test split. GazeCapture releases neither head pose nor 3D face positions, so all methods share the per-frame head pose, camera intrinsics, and 3D face center published with the FAZE preprocessing~\cite{park2019faze}, and this face center serves as the dataset origin. Evaluation covers the $164{,}209$ of $179{,}496$ test frames included in this geometry, which also identifies the subject in each frame.

\subsubsection{MPIIFaceGaze}

Following UniGaze~\cite{qin2026unigaze}, subjects p00 to p09 are used for fine-tuning, and subjects p10 to p14 form the test split with $9{,}055$ images. Inputs are the full camera images with the released calibration and lens distortion. The dataset origin is the released face center, the mean of the four eye corners and two mouth corners, and normalized-face methods use the released head pose.

\subsubsection{EYEDIAP}

Following UniGaze, subjects 1 to 8 are used for fine-tuning and subjects 9 to 16 for testing, using the VGA videos of the continuous screen target (CS) and floating target (FT) sessions. We keep only frames annotated as valid fixations, which excludes blinks and looks away from the target, take the frames whose index is a multiple of $15$, and pool the resulting $7{,}871$ CS and FT frames into one mean error. The dataset origin is the midpoint of the released eyeball centers. Normalized-face methods use the released head rotation, with normalization centers fitted to estimated landmarks.

\subsubsection{Columbia}

Columbia defines no split, so we use the first $37$ subjects in release order for fine-tuning and the remaining $19$ subjects (0038 to 0056) as the test split with $1{,}995$ images. Columbia provides gaze and head angles and a fixed capture distance of $2$ m but no eye positions. We therefore locate the midpoint between the eyes in each image with STAR landmarks and place it at a depth of $2$ m as the dataset origin shared by all methods. Each target lies where the released gaze ray meets the target wall, except for the central target behind the camera, which we place at the camera lens. Camera intrinsics follow the lens setting recorded for each photograph.

\subsubsection{RT-GENE}

RT-GENE defines no subject split, so of its $15$ subjects we use s000 to s009 for fine-tuning and the five subjects s010 to s013 and s016 as the test split with $38{,}850$ frames. Sessions s014 and s015 are second recordings of s002 and s005 and stay with these subjects on the fine-tuning side. We evaluate on the original images rather than the inpainted ones, so the eye-tracking glasses remain visible. Face-input methods read the released narrow face crops, and head-box methods read the released wide crops. The release provides no intrinsics for the crops, so we assign each crop a focal length of $224\,z/s$ pixels. Here $z$ is the released head depth in millimeters, and $s$ is the physical span of the $224$-pixel crop estimated from eye distances, $156.5$ mm for the narrow face crops and $203.8$ mm for the wide crops. RT-GENE provides no 3D gaze targets, so we compare predicted and released directions without origin conversion.

\subsubsection{GFIE}

We use the official test split with $6{,}281$ samples. The dataset origin is the released 3D eye position, obtained from the annotated eye location and the depth measured there. GFIE releases camera calibration and 3D gaze targets but no head pose, so methods that need head pose fit the face model with PnP~\cite{lepetit2009epnp} to landmarks estimated within the SCRFD face box.

\subsection{Reproduction of Compared Methods}
\label{sec:supp:evaluation:methods}

Table~\ref{tab:supp:competitors} summarizes the released weights and input construction of each compared method. Each model constructs its inputs following its training-time conventions. On a dataset included in its training data, it follows the conventions used for that dataset, such as the crop extents of UniGaze on Gaze360 and GazeCapture. Facial landmarks, head pose, normalization centers, and camera intrinsics come from the dataset geometry described above. Where landmarks must be estimated, we run both STAR~\cite{zhou2023star}, trained on 300W~\cite{sagonas2013300w} with $68$ points, and the 2D-FAN-4 model of FAN~\cite{bulat2017facealignment} on the same subject face. For each competing method and dataset, we keep the landmarks that give the lower mean angular error on the common samples, recomputing the common samples after each choice until the choices no longer change. This choice affects only method inputs and method origins, while ground-truth directions and origins remain fixed.

\begin{table*}[!tbp]
  \centering
  \footnotesize
  \caption{Reproduction settings of the compared methods.}
  \label{tab:supp:competitors}
  \begin{tabular*}{\textwidth}{@{\extracolsep{\fill}}l>{\raggedright\arraybackslash}p{0.19\textwidth}>{\raggedright\arraybackslash}p{0.33\textwidth}>{\raggedright\arraybackslash}p{0.23\textwidth}@{}}
    \toprule
    Method & Training data of the weights & Input construction & Gaze origin \\
    \midrule
    GaT baseline~\cite{vuillecard2025gat} & Gaze360 & Same as GaT + ST-WSGE & Midpoint between the eyes \\
    GaT + ST-WSGE & Gaze360, GazeFollow & Head crop on head boxes from its YOLOv5m head detector & Midpoint between the eyes \\
    GeomGaze~\cite{tafasca2023childplay} & Gaze360 and GazeFollow, plus VideoAttentionTarget~\cite{chong2020attended} or ChildPlay~\cite{tafasca2023childplay} for two of the three weights & Square head crop on the head boxes used by FlexGaze & Head box center \\
    3DGazeNet~\cite{ververas2024gazenet} & Gaze360, ETH-XGaze, GazeCapture, MPIIFaceGaze & Face and eye crops under its official rule from $68$ landmarks, released with ETH-XGaze and estimated with STAR or FAN elsewhere, on the official $448$-pixel normalized faces of ETH-XGaze, the released normalized faces of MPIIFaceGaze, the released head crops of Gaze360, and the original images elsewhere & Midpoint between the eyeball centers, or the normalization center on normalized faces \\
    L2CS-Net~\cite{abdelrahman2023l2cs} & Gaze360 & Face crop from its RetinaFace detector, resized to $224$ and then $448$ pixels, without normalization & Midpoint between the eyes \\
    PureGaze~\cite{cheng2022puregaze} & ETH-XGaze & Normalized face in the ETH-XGaze convention & Face center \\
    XGaze baseline~\cite{zhang2020ethxgaze} & ETH-XGaze & Normalized face in the ETH-XGaze convention & Face center \\
    GazeTR-Hybrid~\cite{cheng2022gazetr} & ETH-XGaze & Normalized face in the ETH-XGaze convention & Face center \\
    UniGaze-B/L/H~\cite{qin2026unigaze} & Gaze360, ETH-XGaze, GazeCapture, MPIIFaceGaze, EYEDIAP & Normalized face with the crop extent of its training data on each dataset & Face center \\
    UniGaze-H (XGaze) & ETH-XGaze & Normalized face in the ETH-XGaze convention & Face center \\
    FlexGaze & Generated labels, then the fine-tuning datasets & Head crop through a virtual camera on head boxes from our YOLOv5m head detection & Head box center \\
    \bottomrule
  \end{tabular*}
  \par\vspace{3pt}
  \begin{minipage}{\textwidth}
    The ETH-XGaze normalization uses a virtual camera with a focal length of 960 pixels at a distance of 600 mm and a 224-pixel patch, covering about 140 mm at the face. UniGaze uses its training crop extents of about 166.7 mm on Gaze360 and 153.6 mm on GazeCapture and the ETH-XGaze convention on the other datasets.
  \end{minipage}
\end{table*}

Detection boxes follow the pipeline of each method. L2CS-Net detects faces with its RetinaFace detector and, following its official pipeline, resizes each face crop to $224$ pixels and then to a $448$-pixel input without normalization. Anchor-based face detectors fragment or miss very large faces, as in the close-range faces of Columbia and raw ETH-XGaze images, so we run RetinaFace and SCRFD in two steps. A first pass on a downscaled image estimates the width of the largest face, and if this width exceeds the reliable range of the detector, we downscale the image until the face fits before the final detection. On RT-GENE, L2CS-Net reads each released narrow crop directly as its face crop. GaT + ST-WSGE detects heads with the CrowdHuman-trained YOLOv5m detector used by its official pipeline. It selects the detected box that contains the subject's annotated 2D eye position, in place of the official selection by overlap with the ground-truth head box. On RT-GENE, whose wide crops cut off the top and sides of the head, it reads the head boxes used by FlexGaze. Its midpoint between the eyes comes from released eye annotations where available and from the estimated landmarks otherwise. FlexGaze uses the same YOLOv5m detection and box consolidation as label generation and takes the head box containing the subject's annotated location.

Before scoring, we express all predictions in the original camera frame. Normalized-face methods and FlexGaze predict in a virtual camera, so we apply the inverse of its rotation. L2CS-Net and GaT + ST-WSGE predict relative to the head in their crops, in a frame whose forward axis is the camera ray through the midpoint between the eyes. We therefore rotate their predictions into the camera frame along that ray. We also map the angle and axis conventions of each method to this common camera frame.

\subsection{Constant Mean-Direction Baseline}
\label{sec:supp:evaluation:mean-direction}

The Mean direction rows in Tables~II and~IV predict the same direction for every sample in a column without reading any image. This direction is the normalized sum of the ground-truth unit gaze directions over the common samples of that column, with each direction taken from the dataset origin in the camera frame. Because it uses test labels, this baseline indicates how concentrated each test distribution is rather than serving as a deployable method.

\subsection{Crop Sensitivity Settings}
\label{sec:supp:evaluation:crop-sensitivity}

Fig.~6 compares FlexGaze-B fine-tuned on eight datasets (seed $0$ weights), UniGaze-H~\cite{qin2026unigaze}, and GaT + ST-WSGE~\cite{vuillecard2025gat} on the test sets of Gaze360~\cite{kellnhofer2019gaze360} and ETH-XGaze~\cite{zhang2020ethxgaze}. Each method keeps its own preprocessing pipeline with only its crop window changed, and all inputs are regenerated from the original images. Crop scale is expressed as the ratio of eye distance to crop side, and center shift in units of eye distance. Eye distance is the image-plane distance between the two eye centers, where each eye center is the midpoint of its inner and outer corners. On Gaze360, the eye corners come from STAR~\cite{zhou2023star} landmarks on faces detected by SCRFD~\cite{guo2021scrfd}. On ETH-XGaze, they come from the facial landmarks released with the dataset.

In addition to each method's default crop, the crop scale settings use nine target ratios of eye distance to crop side, $0.15\times1.2^{j}$ for $j=0,\ldots,8$, which range from $0.15$ to about $0.645$. For each method and dataset, we convert every target into a scale factor of the default crop such that the median ratio over the evaluated samples equals the target, and the horizontal axis shows this measured median. Center shifts keep each method's default crop scale and move the crop center by $0.125$, $0.25$, $0.375$, $0.5$, $0.625$, or $0.75$ times the eye distance, again calibrated per method and dataset using the median. The shift direction of each sample is drawn uniformly at random on the circle and shared by all methods, and results are averaged over three direction seeds. Both datasets use the same target values. All curves for a dataset are evaluated on the same samples, namely the common test samples of Table~II for which eye points are available and every method produces a prediction at every setting. These comprise $22{,}831$ samples on Gaze360 and $154{,}158$ on ETH-XGaze.

\section{Additional Results}
\label{sec:supp:results}

This section reports the full comparison, the per-dataset ablation results, and additional crop sensitivity results.

\subsection{Full Comparison on Face-Detected Samples}
\label{sec:supp:results:full-comparison}

Table~\ref{tab:supp-full-comparison} extends Table~II to all released weights that we evaluate under the same protocol and on the same common samples. The additional rows are a median-direction baseline, 3DGazeNet~\cite{ververas2024gazenet}, the GaT baseline without ST-WSGE, three GeomGaze weights trained on different gaze-following datasets, and UniGaze-H trained on ETH-XGaze only. The median-direction baseline predicts the spherical median of the ground-truth directions of each column. Their input construction is given in Sec.~\ref{sec:supp:evaluation:methods} and Table~\ref{tab:supp:competitors}. The training data of 3DGazeNet include all MPIIFaceGaze test samples that we use and possibly the official validation splits of Gaze360 and GazeCapture, as marked in the table. Its errors under its own pipeline, over all frames with predictions, are $4.13^\circ$, $4.49^\circ$, and $10.34^\circ$ on the training, validation, and test splits of Gaze360 and $4.34^\circ$, $4.27^\circ$, and $5.07^\circ$ on those of GazeCapture, which is consistent with training on the validation splits.

\begin{table*}[!tbp]
  \centering
  \footnotesize
  \setlength{\tabcolsep}{0.75pt}
  \renewcommand{\arraystretch}{0.94}
  \continuousresultbackgrounds
  \DeclareRobustCommand{\indom}[1]{%
    \resultcellbackground{\dimexpr\tabcolsep+0.05pt\relax}{\dimexpr\tabcolsep+0.05pt\relax}#1}
  \newlength{\suppfullresultleft}
  \newlength{\suppfullresultmeanleft}
  \newlength{\suppfullresultright}
  \newlength{\suppfullresultwidth}
  \newlength{\suppfullresultshift}
  \DeclareRobustCommand{\suppfullresultheader}[1]{%
    \ifresultmeasure
      \resultmeasurevalue{header}{\begin{tabular}[c]{@{}c@{}}#1\end{tabular}}%
    \fi
    \begin{tabular}[c]{@{}c@{}}#1\end{tabular}}
  \newcommand{\suppfullresultsizes}[1]{%
    \def\resultcolumnname{#1}%
    \setlength{\resultpadding}{\dimexpr\resultwidth{header}/2\relax}%
    \ifdim\resultpadding>\suppfullresultleft\setlength{\suppfullresultleft}{\resultpadding}\fi
    \ifdim\resultpadding>\suppfullresultright\setlength{\suppfullresultright}{\resultpadding}\fi
    \setlength{\resultpadding}{\dimexpr\resultwidth{mean}/2\relax}%
    \ifdim\resultpadding>\suppfullresultmeanleft\setlength{\suppfullresultmeanleft}{\resultpadding}\fi
    \ifdim\resultpadding>\suppfullresultleft\setlength{\suppfullresultleft}{\resultpadding}\fi
    \addtolength{\resultpadding}{\resultwidth{sd}}%
    \ifdim\resultpadding>\suppfullresultright\setlength{\suppfullresultright}{\resultpadding}\fi}
  \renewcommand{\resultcolumnspace}{%
    \ifresultmeasure\else\kern\suppfullresultshift\fi}
  \newcommand{\suppfullresultcolumncell}[2]{%
    \makebox[\suppfullresultwidth][c]{\resultcolumncell{#1}{#2}\resultcolumnspace}}
  \newcolumntype{F}[1]{>{\collectcell{\suppfullresultcolumncell{#1}}}c<{\endcollectcell}}
  \RenewDocumentEnvironment{resulttabular}{m +b}{%
    \stepcounter{resulttable}%
    \resultmeasuretrue
    \sbox{\resultmeasurebox}{\begin{tabular}{#1}#2\end{tabular}}%
    \suppfullresultsizes{Gaze360}\suppfullresultsizes{XGaze}%
    \suppfullresultsizes{GazeCapture}\suppfullresultsizes{MPIIFaceGaze}%
    \suppfullresultsizes{EYEDIAP}\suppfullresultsizes{Columbia}%
    \suppfullresultsizes{RT-GENE}\suppfullresultsizes{GFIE}%
    \addtolength{\suppfullresultleft}{-3pt}%
    \ifdim\suppfullresultleft<\suppfullresultmeanleft
      \setlength{\suppfullresultleft}{\suppfullresultmeanleft}%
    \fi
    \setlength{\suppfullresultwidth}{\dimexpr\suppfullresultleft+\suppfullresultright\relax}%
    \setlength{\suppfullresultshift}{\dimexpr\suppfullresultright-\suppfullresultleft\relax}%
    \resultmeasurefalse
    \begin{tabular}{#1}#2\end{tabular}%
  }{}
  \caption{Comparison of all models measured under the unified evaluation protocol on the face-detected samples (mean angular error in degrees, lower is better).}
  \label{tab:supp-full-comparison}
  \begin{resulttabular}{@{}l!{\hspace{5pt}}N{Params}!{\hspace{5pt}}c!{\hspace{5pt}}F{Gaze360}F{XGaze}F{GazeCapture}F{MPIIFaceGaze}F{EYEDIAP}F{Columbia}F{RT-GENE}F{GFIE}@{}}
    \toprule[0.8pt]
    Method & Params & Preproc.
      & \suppfullresultheader{Gaze360\\{\scriptsize\upshape\color{black}Face (88\%)}} & \suppfullresultheader{XGaze}
      & \suppfullresultheader{Gaze\\Capture} & \suppfullresultheader{MPII\\FaceGaze}
      & \suppfullresultheader{EYEDIAP} & \suppfullresultheader{Columbia}
      & \suppfullresultheader{RT-GENE\\{\scriptsize\upshape\color{black}Original}} & \suppfullresultheader{GFIE\\{\scriptsize\upshape\color{black}Face (86\%)}} \\
    \midrule[0.6pt]
    \multicolumn{11}{l}{\textbf{Oracle baselines with constant predictions}} \\
    Mean direction & --- & ---
      & 51.48 & 42.64 & 7.78 & 11.75 & 23.83 & 12.05 & 27.18 & 36.89 \\
    Median direction & --- & ---
      & 51.30 & 42.64 & 7.78 & 11.75 & 23.73 & 12.05 & 27.16 & 36.31 \\
    \midrule[0.4pt]
    \multicolumn{11}{l}{\textbf{Baselines with head-crop input}} \\
    GaT baseline~\cite{vuillecard2025gat} & 27.9M & H
      & \indom{12.24} & 16.33 & 8.37 & 9.22 & 8.50 & 8.57 & 9.91\pmark{\,c} & 19.26 \\
    \methodextra{GaT + ST-WSGE}{+GazeFollow} & 27.9M & H
      & \indom{11.90} & 14.36 & 6.95 & 6.79 & 8.56 & 8.07 & 9.54\pmark{\,c} & 15.06 \\
    \methodextra{GeomGaze~\cite{tafasca2023childplay}$^{c}$}{+GazeFollow} & 11.4M & H
      & \indom{29.66} & 45.65 & 34.40 & 27.98 & 28.99 & 47.50 & 30.65 & 13.27 \\
    \methodextra{GeomGaze$^{c}$}{+VideoAttentionTarget} & 11.4M & H
      & \indom{22.40} & 29.48 & 23.01 & 17.56 & 19.55 & 25.38 & 22.93 & 16.69 \\
    \methodextra{GeomGaze$^{c}$}{+ChildPlay} & 11.4M & H
      & \indom{22.72} & 37.63 & 30.69 & 21.33 & 27.87 & 28.94 & 27.91 & 14.93 \\
    \midrule[0.4pt]
    \multicolumn{11}{l}{\textbf{Baseline with face-and-eye-crop input}} \\
    3DGazeNet~\cite{ververas2024gazenet} & 30.2M & F,L
      & \indom{\textbf{10.89}\pmark{\,a}} & \indom{5.39} & \indom{4.06\pmark{\,a}} & \indom{2.11\pmark{\,b}} & 9.60 & 4.41 & 39.81 & 28.54 \\
    \midrule[0.4pt]
    \multicolumn{11}{l}{\textbf{Baselines with face-crop input}} \\
    L2CS-Net~\cite{abdelrahman2023l2cs} & 23.9M & F
      & \indom{16.66} & 17.52 & 10.20 & 9.01 & 11.62 & 8.20 & 30.22 & 38.32 \\
    PureGaze~\cite{cheng2022puregaze} & 23.5M & F,L,P,K
      & 33.32 & \indom{4.62} & 7.69 & 9.21 & 8.31 & 7.65 & 18.41 & 57.79 \\
    XGaze baseline~\cite{zhang2020ethxgaze} & 23.5M & F,L,P,K
      & 34.97 & \indom{4.44} & 8.50 & 9.07 & 10.03 & 9.19 & 17.24 & 60.90 \\
    GazeTR-Hybrid~\cite{cheng2022gazetr} & 11.4M & F,L,P,K
      & 34.15 & \indom{\textbf{4.29}} & 10.66 & 10.06 & 13.07 & 9.53 & 20.15 & 49.43 \\
    UniGaze-H (XGaze)~\cite{qin2026unigaze} & 630.8M & F,L,P,K
      & 20.06 & \indom{5.13} & 5.84 & 7.00 & 6.47 & 6.15 & 10.59 & 34.55 \\
    UniGaze-B & 85.8M & F,L,P,K
      & \indom{16.19} & \indom{5.73} & \indom{3.83} & \indom{5.80} & \indom{5.16} & 4.56 & 11.00 & 31.96 \\
    UniGaze-L & 303.3M & F,L,P,K
      & \indom{15.77} & \indom{5.61} & \indom{3.55} & \indom{5.30} & \indom{4.98} & \textbf{4.22} & 15.97 & 30.47 \\
    UniGaze-H & 630.8M & F,L,P,K
      & \indom{16.12} & \indom{5.40} & \indom{3.50} & \indom{5.09} & \indom{\textbf{4.83}} & 4.46 & 26.87 & 30.92 \\
    \midrule[0.6pt]
    \multicolumn{11}{l}{\textbf{Ours, head-crop input, fine-tuned (8 sets)}} \\
    FlexGaze-S & 21.8M & H,K
      & \indom{\cellstd{11.11}{0.08}} & \indom{\cellstd{5.97}{0.04}} & \indom{\cellstd{3.68}{0.03}} & \indom{\cellstd{5.35}{0.16}}
      & \indom{\cellstd{5.86}{0.10}} & \indom{\cellstd{5.01}{0.11}} & \indom{\cellstd{6.86}{0.20}} & \indom{\cellstd{12.17}{0.24}} \\
    FlexGaze-B & 86.0M & H,K
      & \indom{\cellstd{10.95}{0.10}} & \indom{\cellstd{5.31}{0.06}} & \indom{\cellstd{\textbf{3.46}}{0.03}} & \indom{\cellstd{\textbf{4.92}}{0.25}}
      & \indom{\cellstd{5.39}{0.01}} & \indom{\cellstd{4.42}{0.15}} & \indom{\cellstd{\textbf{6.20}}{0.10}} & \indom{\cellstd{\textbf{11.09}}{0.54}} \\
    \midrule[0.4pt]
    \multicolumn{11}{l}{\textbf{Ours, head-crop input, fine-tuned (5 sets shared with UniGaze)}} \\
    FlexGaze-S & 21.8M & H,K
      & \indom{\cellstd{11.17}{0.07}} & \indom{\cellstd{5.98}{0.04}} & \indom{\cellstd{3.67}{0.05}} & \indom{\cellstd{5.54}{0.15}}
      & \indom{\cellstd{5.90}{0.06}} & \cellstd{5.54}{0.42} & \cellstd{10.41}{0.66} & \cellstd{17.70}{1.10} \\
    FlexGaze-B & 86.0M & H,K
      & \indom{\cellstd{10.90}{0.15}} & \indom{\cellstd{5.30}{0.03}} & \indom{\cellstd{3.46}{0.01}} & \indom{\cellstd{4.93}{0.21}}
      & \indom{\cellstd{5.33}{0.05}} & \cellstd{4.52}{0.14} & \cellstd{10.21}{0.25} & \cellstd{14.34}{0.61} \\
    \midrule[0.4pt]
    \multicolumn{11}{l}{\textbf{Ours, head-crop input, pretrained (weak labels only)}} \\
    FlexGaze-S & 21.8M & H,K
      & \cellstd{12.69}{0.03} & \cellstd{12.90}{0.07} & \cellstd{10.59}{0.26} & \cellstd{9.97}{0.26}
      & \cellstd{8.35}{0.27} & \cellstd{9.32}{0.30} & \cellstd{11.62}{0.34} & \cellstd{12.19}{0.15} \\
    FlexGaze-B & 86.0M & H,K
      & \cellstd{12.47}{0.11} & \cellstd{13.54}{0.16} & \cellstd{12.28}{0.92} & \cellstd{11.67}{0.99}
      & \cellstd{8.71}{0.16} & \cellstd{8.00}{0.44} & \cellstd{12.24}{0.65} & \cellstd{12.69}{0.25} \\
    \bottomrule[0.8pt]
  \end{resulttabular}
  \par\vspace{3pt}
  {\footnotesize
  \begin{tabular}{@{}l@{\enspace}>{\raggedright\arraybackslash}p{0.84\textwidth}@{}}
  Params & Inference parameters, excluding preprocessing models. \\
  Preproc. & Input preparation requirements: H (head box), F (face box), L (facial landmarks), P (head pose), K (camera intrinsics). \\
  \colorbox{gray!10}{\strut in-domain} & The model was trained on this dataset. \\
  {\scriptsize\color{black}+Dataset} & The model was additionally trained with gaze-related labels from this dataset. Table~\ref{tab:supp:competitors} lists the full training data. \\
  UniGaze-H (XGaze) & UniGaze-H weights trained on ETH-XGaze only. \\
  mean\stdsuffix{SD} & FlexGaze rows give the mean and sample SD over seeds $0$, $1$, and $2$. \\
  \textbf{bold} & Lowest error in each column among the compared methods and the eight-dataset fine-tuned models. Results marked $b$ are not considered for bold. \\
  $^{a}$ & The model may have been trained on the official validation split of this dataset. \\
  $^{b}$ & All test samples we use for this dataset are included in the training data of the model. \\
  $^{c}$ & The model has no head detection of its own (GeomGaze, all datasets), or its head detector finds the subject in only a small share of the images (GaT, RT-GENE), so it reads the head boxes used by FlexGaze. \\
  \addlinespace[2pt]
  Not evaluated & OmniGaze~\cite{qu2025omnigaze} could not be evaluated because its model weights could not be obtained and its test splits are insufficiently specified. GazeTR-Pure~\cite{cheng2022gazetr} could not be evaluated because only the GazeTR-Hybrid weights are released. \\
  \end{tabular}
  \par
  }
\end{table*}

\begin{table*}[!t]
  \centering
  \footnotesize
  \setlength{\tabcolsep}{0.4pt}
  \newlength{\suppablremovedwidth}
  \newlength{\suppablremovedcandidate}
  \settowidth{\suppablremovedwidth}{\scriptsize DINOv3 init.}
  \settowidth{\suppablremovedcandidate}{\scriptsize Separate heads}
  \ifdim\suppablremovedcandidate>\suppablremovedwidth\setlength{\suppablremovedwidth}{\suppablremovedcandidate}\fi
  \settowidth{\suppablremovedcandidate}{\scriptsize FlexGaze init.}
  \ifdim\suppablremovedcandidate>\suppablremovedwidth\setlength{\suppablremovedwidth}{\suppablremovedcandidate}\fi
  \DeclareRobustCommand{\ablationreplace}[2]{$-$ \makebox[\suppablremovedwidth][l]{#1} $+$ #2}
  \continuousresultbackgrounds
  \DeclareRobustCommand{\indom}[1]{%
    \resultcellbackground{\dimexpr\tabcolsep+0.05pt\relax}{\dimexpr\tabcolsep+0.05pt\relax}#1}
  \newlength{\suppablresultleft}
  \newlength{\suppablresultmeanleft}
  \newlength{\suppablresultright}
  \newlength{\suppablresultwidth}
  \newlength{\suppablresultshift}
  \DeclareRobustCommand{\suppablresultheader}[1]{%
    \ifresultmeasure
      \resultmeasurevalue{header}{\begin{tabular}[c]{@{}c@{}}#1\end{tabular}}%
    \fi
    \begin{tabular}[c]{@{}c@{}}#1\end{tabular}}
  \newcommand{\suppablresultsizes}[1]{%
    \def\resultcolumnname{#1}%
    \setlength{\resultpadding}{\dimexpr\resultwidth{header}/2\relax}%
    \ifdim\resultpadding>\suppablresultleft\setlength{\suppablresultleft}{\resultpadding}\fi
    \ifdim\resultpadding>\suppablresultright\setlength{\suppablresultright}{\resultpadding}\fi
    \setlength{\resultpadding}{\dimexpr\resultwidth{mean}/2\relax}%
    \ifdim\resultpadding>\suppablresultmeanleft\setlength{\suppablresultmeanleft}{\resultpadding}\fi
    \ifdim\resultpadding>\suppablresultleft\setlength{\suppablresultleft}{\resultpadding}\fi
    \addtolength{\resultpadding}{\resultwidth{sd}}%
    \ifdim\resultpadding>\suppablresultright\setlength{\suppablresultright}{\resultpadding}\fi}
  \renewcommand{\resultcolumnspace}{%
    \ifresultmeasure\else\kern\suppablresultshift\fi}
  \newcommand{\suppablresultcolumncell}[2]{%
    \makebox[\suppablresultwidth][c]{\resultcolumncell{#1}{#2}\resultcolumnspace}}
  \newcolumntype{Q}[1]{>{\collectcell{\suppablresultcolumncell{#1}}}c<{\endcollectcell}}
  \RenewDocumentEnvironment{resulttabular}{m +b}{%
    \stepcounter{resulttable}%
    \resultmeasuretrue
    \sbox{\resultmeasurebox}{\begin{tabular}{#1}#2\end{tabular}}%
    \suppablresultsizes{Gaze360}\suppablresultsizes{XGaze}%
    \suppablresultsizes{GazeCapture}\suppablresultsizes{MPIIFaceGaze}%
    \suppablresultsizes{EYEDIAP}\suppablresultsizes{Columbia}%
    \suppablresultsizes{RT-GENE}\suppablresultsizes{GFIE}%
    \addtolength{\suppablresultleft}{-3pt}%
    \ifdim\suppablresultleft<\suppablresultmeanleft
      \setlength{\suppablresultleft}{\suppablresultmeanleft}%
    \fi
    \setlength{\suppablresultwidth}{\dimexpr\suppablresultleft+\suppablresultright\relax}%
    \setlength{\suppablresultshift}{\dimexpr\suppablresultright-\suppablresultleft\relax}%
    \resultmeasurefalse
    \begin{tabular}{#1}#2\end{tabular}%
  }{}
  \caption{Per-dataset results of the FlexGaze-B ablations (mean angular error, $^\circ$).}
  \label{tab:supp-ablation-per-dataset}
  \begin{resulttabular}{@{}l!{\hspace{8pt}}Q{Gaze360}Q{XGaze}Q{GazeCapture}Q{MPIIFaceGaze}Q{EYEDIAP}Q{Columbia}Q{RT-GENE}Q{GFIE}@{}}
    \toprule[0.8pt]
    Setting
      & \suppablresultheader{Gaze360\\{\scriptsize\upshape\color{black}Face (88\%)}} & \suppablresultheader{XGaze}
      & \suppablresultheader{Gaze\\Capture} & \suppablresultheader{MPII\\FaceGaze}
      & \suppablresultheader{EYEDIAP} & \suppablresultheader{Columbia}
      & \suppablresultheader{RT-GENE\\{\scriptsize\upshape\color{black}Original}} & \suppablresultheader{GFIE\\{\scriptsize\upshape\color{black}Face (86\%)}} \\
    \midrule[0.8pt]
    \multicolumn{9}{@{}l@{}}{\textbf{Pretrained (weak labels only)}} \\
    Full recipe
      & \cellstd{12.47}{0.11}
      & \cellstd{\textbf{13.54}}{0.16}
      & \cellstd{12.28}{0.92}
      & \cellstd{11.67}{0.99}
      & \cellstd{8.71}{0.16}
      & \cellstd{8.00}{0.44}
      & \cellstd{\textbf{12.24}}{0.65}
      & \cellstd{12.69}{0.25}
      \\
    {\scriptsize $-$ Pose-unavailable samples}
      & \cellstd{\textbf{12.41}}{0.09}
      & \cellstd{13.65}{0.31}
      & \cellstd{12.00}{1.34}
      & \cellstd{11.11}{1.09}
      & \cellstd{\textbf{8.40}}{0.16}
      & \cellstd{\textbf{7.78}}{0.21}
      & \cellstd{12.26}{0.65}
      & \cellstd{\textbf{12.37}}{0.14}
      \\
    {\scriptsize $-$ Crop jitter}
      & \cellstd{12.79}{0.21}
      & \cellstd{13.99}{0.25}
      & \cellstd{12.48}{1.10}
      & \cellstd{11.71}{1.43}
      & \cellstd{8.73}{0.23}
      & \cellstd{8.47}{0.24}
      & \cellstd{13.11}{1.13}
      & \cellstd{13.25}{0.44}
      \\
    {\scriptsize \ablationreplace{DINOv3 init.}{Random init.}}
      & \cellstd{18.66}{0.10}
      & \cellstd{25.73}{0.63}
      & \cellstd{\textbf{10.46}}{0.12}
      & \cellstd{\textbf{10.16}}{0.34}
      & \cellstd{15.58}{0.32}
      & \cellstd{18.25}{0.78}
      & \cellstd{14.68}{0.15}
      & \cellstd{16.56}{0.12}
      \\
    \midrule[0.4pt]
    \multicolumn{9}{@{}l@{}}{\textbf{Fine-tuned (5 sets shared with UniGaze)}} \\
    Full recipe
      & \indom{\cellstd{10.90}{0.15}}
      & \indom{\cellstd{5.30}{0.03}}
      & \indom{\cellstd{3.46}{0.01}}
      & \indom{\cellstd{4.93}{0.21}}
      & \indom{\cellstd{5.33}{0.05}}
      & \cellstd{\textbf{4.52}}{0.14}
      & \cellstd{10.21}{0.25}
      & \cellstd{14.34}{0.61}
      \\
    {\scriptsize \ablationreplace{Separate heads}{Shared head}}
      & \indom{\cellstd{\textbf{10.56}}{0.02}}
      & \indom{\cellstd{\textbf{5.28}}{0.02}}
      & \indom{\cellstd{3.47}{0.03}}
      & \indom{\cellstd{4.87}{0.21}}
      & \indom{\cellstd{\textbf{5.15}}{0.02}}
      & \cellstd{6.40}{0.35}
      & \cellstd{12.36}{0.26}
      & \cellstd{\textbf{12.94}}{0.31}
      \\
    {\scriptsize $-$ Pseudo-labeled data replay}
      & \indom{\cellstd{11.00}{0.13}}
      & \indom{\cellstd{5.40}{0.03}}
      & \indom{\cellstd{3.49}{0.02}}
      & \indom{\cellstd{4.93}{0.17}}
      & \indom{\cellstd{5.60}{0.18}}
      & \cellstd{4.66}{0.19}
      & \cellstd{10.48}{1.28}
      & \cellstd{15.58}{1.67}
      \\
    {\scriptsize $-$ Crop jitter}
      & \indom{\cellstd{11.03}{0.06}}
      & \indom{\cellstd{5.47}{0.04}}
      & \indom{\cellstd{\textbf{3.39}}{0.02}}
      & \indom{\cellstd{\textbf{4.81}}{0.14}}
      & \indom{\cellstd{5.47}{0.07}}
      & \cellstd{4.99}{0.10}
      & \cellstd{10.63}{0.52}
      & \cellstd{14.43}{0.60}
      \\
    {\scriptsize \ablationreplace{FlexGaze init.}{DINOv3 init.}}
      & \indom{\cellstd{11.08}{0.13}}
      & \indom{\cellstd{5.30}{0.05}}
      & \indom{\cellstd{3.48}{0.04}}
      & \indom{\cellstd{5.13}{0.10}}
      & \indom{\cellstd{5.37}{0.06}}
      & \cellstd{5.18}{0.44}
      & \cellstd{\textbf{10.17}}{0.96}
      & \cellstd{14.70}{1.51}
      \\
    \bottomrule[0.8pt]
  \end{resulttabular}
  \par\vspace{2pt}
  {\scriptsize\raggedright

  Rows correspond to Table~V, whose averages are computed from these errors. Evaluation samples and scoring follow Table~II. Each cell gives the mean and SD over seeds $0$, $1$, and $2$. Gray marks datasets used for ground-truth fine-tuning, and bold marks the lowest error in each column within each training stage.\par}
\end{table*}

\begin{figure*}[!tbp]
  \centering
  \includegraphics[width=\textwidth]{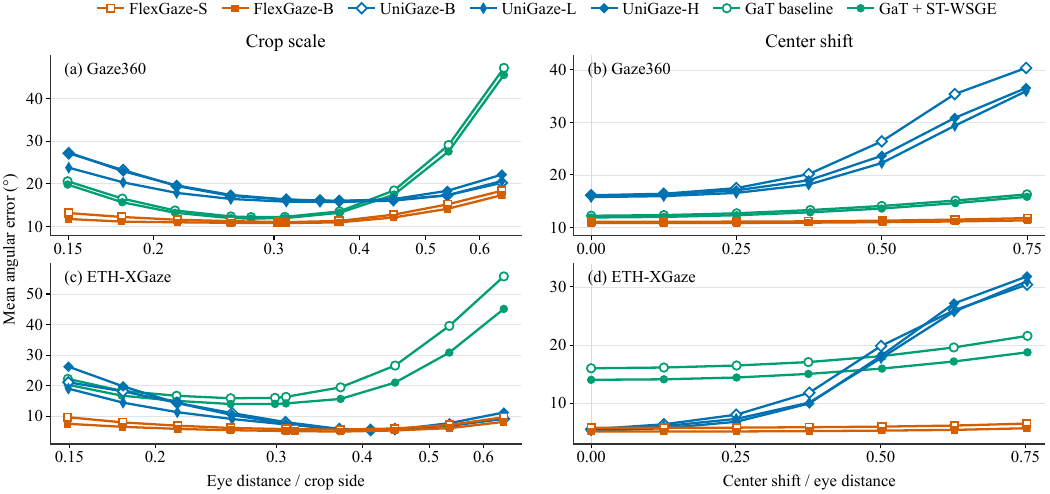}
  \caption{Crop sensitivity of all evaluated models. FlexGaze-S and FlexGaze-B fine-tuned on eight datasets, UniGaze-B/L/H, the GaT baseline, and GaT + ST-WSGE are compared in the setting of Fig.~6, with FlexGaze from training seed $0$. Rows show Gaze360 and ETH-XGaze, and columns show crop scale and center shift with the axes and settings of Fig.~6. Larger ratios of eye distance to crop side indicate tighter crops.}
  \label{fig:supp-sensitivity-all-models}
\end{figure*}

\begin{figure*}[!tbp]
  \centering
  \includegraphics[width=\textwidth]{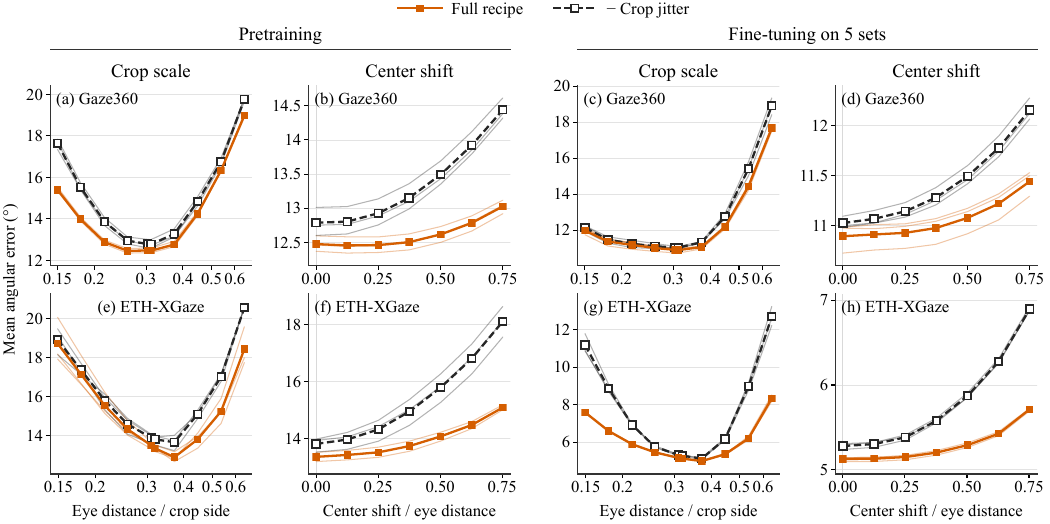}
  \caption{Crop sensitivity of FlexGaze-B with and without crop jitter. Vermilion solid lines with filled squares show the full recipe, and dark dashed lines with open squares show the model trained without crop jitter. Panels (a,b,e,f) show pretraining and (c,d,g,h) five-dataset fine-tuning, with Gaze360 in the top row and ETH-XGaze in the bottom row. Crop jitter is disabled separately at each stage, and both fine-tuned models start from the weights pretrained with the full recipe. Camera roll augmentation is retained in both settings. Thick lines show the mean over training seeds $0$, $1$, and $2$, and thin lines show individual training seeds. Axes and settings follow Fig.~6.}
  \label{fig:supp-crop-jitter-control}
\end{figure*}

\begin{figure}[!tbp]
  \centering
  \includegraphics[width=\columnwidth]{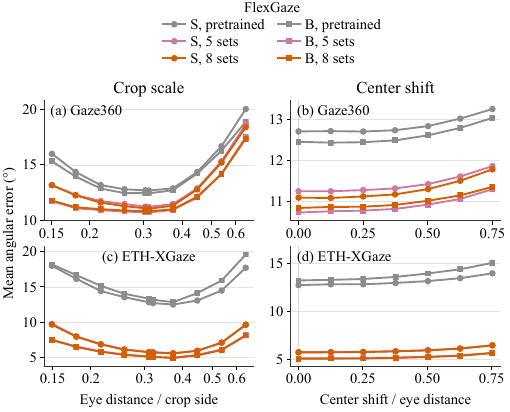}
  \caption{Crop sensitivity of FlexGaze after each training stage. Gray, purple, and vermilion show pretraining, five-dataset fine-tuning, and eight-dataset fine-tuning, with circles for FlexGaze-S and squares for FlexGaze-B, all from training seed $0$. Axes and settings follow Fig.~6. On ETH-XGaze, the curves after five-dataset and eight-dataset fine-tuning nearly coincide.}
  \label{fig:supp-sensitivity-stages}
\end{figure}

\subsection{Per-Dataset Ablation Results}
\label{sec:supp:results:ablation-per-dataset}

Table~\ref{tab:supp-ablation-per-dataset} lists the per-dataset errors of the ablation settings in Table~V, from which its three averages are computed. Settings, evaluation samples, and scoring are identical to those of Table~V. Each ablation changes one setting relative to the full recipe of its training stage. The DINOv3 initialization row therefore keeps pseudo-labeled data replay, and the row without replay keeps the pretrained initialization. These two ablations assess the contribution of generated labels at each stage while retaining them at the other stage. Removing either component increases all three average errors in Table~V.

\subsection{Crop Sensitivity: Additional Results}
\label{sec:supp:results:crop-sensitivity}

Figs.~\ref{fig:supp-sensitivity-all-models}, \ref{fig:supp-crop-jitter-control}, and~\ref{fig:supp-sensitivity-stages} extend the crop sensitivity experiment of Fig.~6 with the settings of Sec.~\ref{sec:supp:evaluation:crop-sensitivity}.

Fig.~\ref{fig:supp-sensitivity-all-models} adds FlexGaze-S, UniGaze-B and UniGaze-L, and the GaT baseline to the models of Fig.~6. The curves of both FlexGaze models stay low and comparatively flat. The errors of all three UniGaze models rise steeply under center shifts and for looser crops, whereas the errors of both GaT models rise mainly for tighter crops.

Fig.~\ref{fig:supp-crop-jitter-control} compares FlexGaze-B trained with the full recipe and without crop jitter. Crop jitter is disabled separately at each stage, and both five-dataset fine-tuned models start from the weights pretrained with the full recipe. Without crop jitter, the errors rise faster under center shifts at both training stages and on both datasets, consistently across the three training seeds. For crop scale, removing crop jitter generally increases the error for tighter crops. For looser crops, the increase is large for the pretrained model evaluated on Gaze360 and for the fine-tuned model evaluated on ETH-XGaze, and small or absent in the other two cases. At the default crop, the mean errors differ by less than $0.5^\circ$.

Fig.~\ref{fig:supp-sensitivity-stages} compares FlexGaze after pretraining, five-dataset fine-tuning, and eight-dataset fine-tuning. Fine-tuning lowers the errors over most of the tested range. On Gaze360, the errors of the three stages rise by similar amounts under center shifts and for tighter crops, whereas for looser crops the errors of the fine-tuned models rise less. On ETH-XGaze, the errors of the fine-tuned models rise less than those of the pretrained models under every crop change.

\FloatBarrier
\section{Data and Ethics Statement}
\label{sec:supp:ethics}

FlexGaze is pretrained only on images from public datasets, which we use under the terms of their respective licenses. Fine-tuning and evaluation use public gaze benchmarks. We did not collect new data from human subjects. The images in these datasets contain faces and other personal information. We release the unfiltered generated annotations and associated metadata but not the source images, so users must obtain the images from their original sources and comply with the corresponding licenses. Gaze estimation can be misused, for example to monitor people's attention without their consent. We ask users to comply with applicable regulations and obtain any required consent when using our models and annotations.

{\small
\bibliographystyle{ieee}
\bibliography{references,appendix-references}
}